\documentclass[journal,twoside,web]{ieeecolor}
\usepackage{etoolbox}
\makeatletter
\@ifundefined{color@begingroup}%
{\let\color@begingroup\relax
\let\color@endgroup\relax}{}%
\def\fix@ieeecolor@hbox#1{%
\hbox{\color@begingroup#1\color@endgroup}}
\patchcmd\@makecaption{\hbox}{\fix@ieeecolor@hbox}{}{\FAILED}
\patchcmd\@makecaption{\hbox}{\fix@ieeecolor@hbox}{}{\FAILED}

\usepackage{tmi}
\usepackage{cite}
\usepackage[normalem]{ulem}
\usepackage{utfsym}
\useunder{\uline}{\ul}{}
\usepackage{amsmath,amssymb,amsfonts}
\usepackage{algorithmic}
\usepackage{algorithm}
\usepackage{graphicx}
\usepackage{textcomp}
\usepackage{multirow}
\usepackage{booktabs}
\usepackage{makecell}

\definecolor{DarkGreen}{RGB}{1,50,32}
\def\BibTeX{{\rm B\kern-.05em{\sc i\kern-.025em b}\kern-.08em
    T\kern-.1667em\lower.7ex\hbox{E}\kern-.125emX}}
\begin{document}
\title{HisynSeg: Weakly-Supervised Histopathological Image Segmentation via Image-Mixing Synthesis and Consistency Regularization}
\author{Zijie Fang, Yifeng Wang, Peizhang Xie, Zhi Wang and Yongbing Zhang 
\thanks{This work was supported in part by the National Natural Science Foundation of China under 62031023 \& 62331011; in part by the Shenzhen Science and Technology Project under GXWD20220818170353009. (Corresponding author: Yongbing Zhang.)}
\thanks{Zijie Fang and Zhi Wang are with the Tsinghua Shenzhen International Graduate School, Tsinghua University, Shenzhen, SZ 518000 CHN (e-mail: vison307@gmail.com, wangzhi@sz.tsinghua.edu.cn).}
\thanks{Yifeng Wang is with the School of Science, Harbin Institute of Technology (Shenzhen), Shenzhen, SZ 518000 CHN (e-mail: wangyifeng@stu.hit.edu.cn).}
\thanks{Peizhang Xie and Yongbing Zhang are with the School of Computer Science and Technology, Harbin Institute of Technology (Shenzhen), Shenzhen, SZ 518000 CHN (e-mail: davidxie03@163.com, ybzhang08@hit.edu.cn).}
}

\maketitle

\begin{abstract}
Tissue semantic segmentation is one of the key tasks in computational pathology. To avoid the expensive and laborious acquisition of pixel-level annotations, a wide range of studies attempt to adopt the class activation map (CAM), a weakly-supervised learning scheme, to achieve pixel-level tissue segmentation. However, CAM-based methods are prone to suffer from under-activation and over-activation issues, leading to poor segmentation performance. To address this problem, we propose a novel weakly-supervised semantic segmentation framework for histopathological images based on image-mixing synthesis and consistency regularization, dubbed HisynSeg. Specifically, synthesized histopathological images with pixel-level masks are generated for fully-supervised model training, where two synthesis strategies are proposed based on Mosaic transformation and Bézier mask generation. Besides, an image filtering module is developed to guarantee the authenticity of the synthesized images. In order to further avoid the model overfitting to the occasional synthesis artifacts, we additionally propose a novel self-supervised consistency regularization, which enables the real images without segmentation masks to supervise the training of the segmentation model. By integrating the proposed techniques, the HisynSeg framework successfully transforms the weakly-supervised semantic segmentation problem into a fully-supervised one, greatly improving the segmentation accuracy. Experimental results on three datasets prove that the proposed method achieves a state-of-the-art performance. Code is available at https://github.com/Vison307/HisynSeg.
\end{abstract}


\section{Introduction}
\label{sec:introduction}
\IEEEPARstart{C}{ancer} is one of the major leading causes of global population death \cite{siegelcancer}. Among all cancer diagnosis methods, pathological diagnosis is the gold standard. Since the U.S. Food and Drug Administration approved the use of whole slide image (WSI) scanners for the initial pathological diagnosis of cancer in 2017 \cite{evans2018us}, the digital observation and preservation of tissue slides have become a reality. Along with the rapid development of artificial intelligence, utilizing deep learning for automatic pathological diagnosis with digital WSIs has become a new trend, giving birth to a new cross-discipline named computational pathology \cite{echle2021deep}. In computational pathology, tissue semantic segmentation is one of the key tasks due to its ability to differentiate various types of tissues. Different tissue types, such as tumor, stroma, lymphocyte, and normal tissues, form the tumor microenvironment (TME), which plays a key role in the generation and development of cancer, and the therapeutic efficacy and prognosis of cancer patients are closely related to the TME \cite{abduljabbar2020geospatial}.

To enable automated tissue semantic segmentation, a large number of researchers have designed a series of deep-learning models. For example, Ronneberger et al. \cite{ronneberger2015u} developed a U-shaped network called U-Net by proposing skip connections between encoders and decoders to enhance the segmentation quality of boundaries. After that, many U-Net-based segmentation networks have been proposed \cite{zhou2018unet++, cao2022swin}. Although these deep learning models have achieved satisfactory tissue segmentation results and provided great opportunities for automated TME analysis, all of them are based on fully-supervised learning strategies. In fully-supervised segmentation, pixel-level segmentation masks are required for model training. However, obtaining dense pixel-level masks is expensive and tedious. Additionally, due to the specialized nature of histopathological images, only trained clinicians or pathologists can perform the annotation task, which further increases the difficulty of acquiring fine annotations. A study \cite{han2022multi} has shown that the average time required to perform pixel-level annotations for 100 histopathological images under a $224\times 224$ size is more than 200 minutes. Besides, pixel-level fine annotations suffer from variations among different annotators due to subjectivity, leading to unavoidable noises in annotations.
\begin{figure}[!t]
    \centering
    \includegraphics[width=0.75\linewidth]{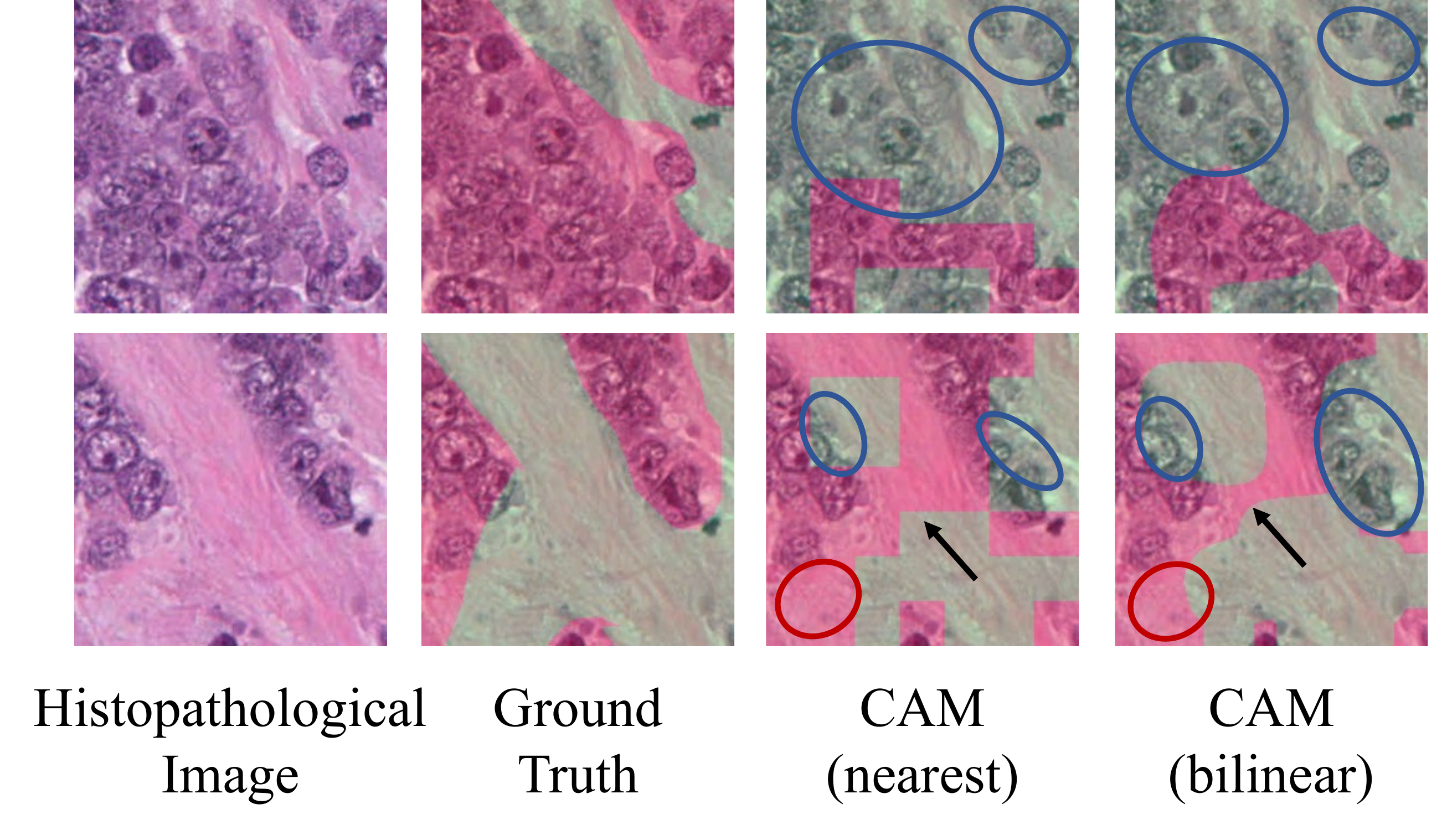}
    \caption{Examples of histopathological images and pseudo-masks generated by CAM. For comparison, ground-truth masks are also provided. Nearest and bilinear represent the utilized interpolation method. Blue and red circles highlight the under-activated and over-activated regions, respectively. Black arrows indicate the introduced noise caused by interpolation. Red pixels represent tumor epithelial and green pixels stand for necrosis. These images are from the BCSS dataset \cite{amgad2019structured}.}
    \label{fig:r2-1}
\end{figure}

In order to solve the above problems, some studies try to achieve fine-grained semantic segmentation based on weakly-supervised learning frameworks using weak labels such as classification labels \cite{chan2019histosegnet}, point annotations \cite{bearman2016s}, and scribbles \cite{lin2016scribblesup}. Among all forms of weak labels, classification labels are the easiest to acquire. Compared with pixel-level annotations, which take 200 minutes to annotate 100 images, assigning classification labels to 100 histopathological images takes only about 6 minutes. Besides, the inconsistency among different annotators can be greatly mitigated \cite{han2022multi}. To this end, using classification labels for tissue semantic segmentation has received much attention. To obtain pixel-level semantic information from image-level classification labels, most researchers adopt the class activation map (CAM) \cite{zhou2016learning} strategy and develop a two-stage framework for segmentation. In the first stage, a classification model is trained using the image-level labels. Then, taking advantage of the trained classification model,  whose feature layers learn the semantic location information of the objects, low-resolution CAMs are generated and up-sampled back to the image size to serve as the pseudo-masks. In the second stage, images with the pseudo-masks are fed into the segmentation model as training samples, ultimately achieving pixel-level segmentation.

However, as shown in Fig.~\ref{fig:r2-1}, CAM-based methods are often accused of their under-activation (the boundaries of the foreground objects are segmented as the background) and over-activation (regions in the background are misclassified as the foreground) issues because merely focusing on the most discriminative regions can already obtain satisfactory classification results \cite{wang2020self}. On the contrary, the segmentation task needs to find out the whole object. In addition, the up-sampling of CAM in the pseudo-mask generation process inevitably introduces noises \cite{zhonghamil}. Although bilinear interpolation can alleviate this problem, inaccuracy still exists. All the above problems lead to the performance degradation of segmentation models. Therefore, obtaining accurate pixel-level masks from the weak classification labels remains a major challenge for weakly-supervised tissue semantic segmentation. 

\begin{figure}[!t]
    \centering
    \includegraphics[width=0.85\linewidth]{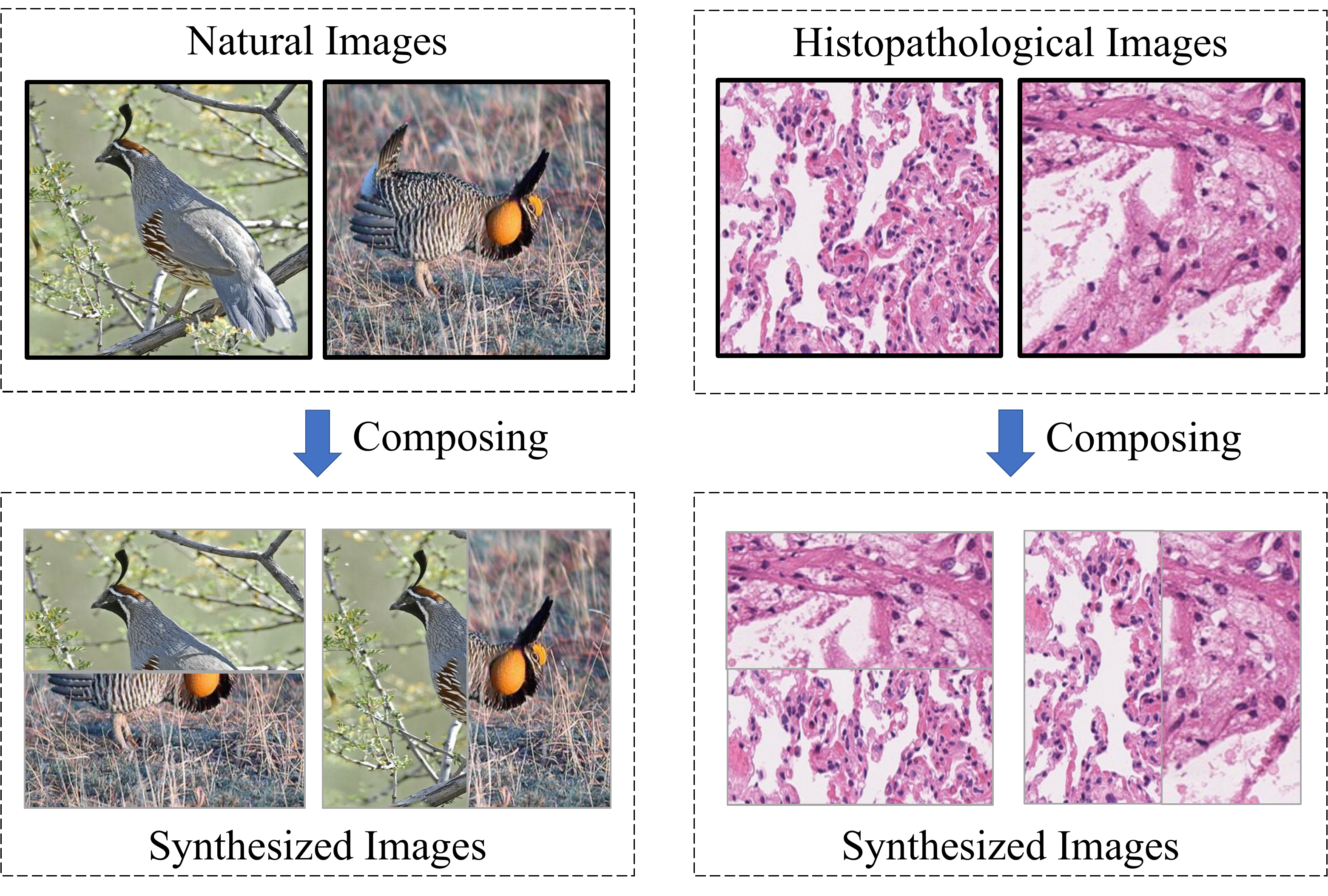}
    \caption{A comparison of image synthesis between natural and histopathological images. Compared with natural images, histopathological images are easier to synthesize due to more uniform colors, milder foreground and background differences, and homogeneity in image content. The natural images are taken from the ImageNet dataset \cite{deng2009imagenet}. The histopathological images are from the WSSS4LUAD dataset \cite{hanwsssluad}.}
    \label{fig:1}
\end{figure}

Different from natural images, which are often captured in various environments with huge differences in image contents, histopathological images are usually taken in a laboratory environment, and the image content is a random repetition of cells, tissues, and background white regions \cite{han2022multi}. Therefore, homogeneity exists both within and between histopathological images, as shown in Fig. \ref{fig:1}. 
The homogeneity property makes the synthesis of histopathological images much easier than natural images. Besides, tissues within the same category tend to cluster together in the TME, making it common for a histopathological image to have only one type of tissue. For this reason, synthesized histopathological images with pixel-level segmentation masks can be generated using histopathological images with a single tissue type. Since the segmentation masks of the images with a single tissue type are definitely accurate, the pixel-level segmentation masks of the synthesized images are also accurate, which can be served as the supervision to train the segmentation models, creating a new paradigm that enables fully-supervised learning models to solve the weakly-supervised segmentation task.

To this end, based on our previous work \cite{fang2023weakly}, this paper proposes a novel image-mixing synthesis empowered weakly-supervised semantic segmentation (WSSS) framework, named HisynSeg, for histopathological images. Compared to our previous work, this paper designs a new image-mixing synthesis strategy based on Bézier mask generation and develops an image filtering module to ensure the authenticity of the synthesized images. In order to further alleviate the model overfitting the occasional artifacts that are inconsistent with the real histopathological images, this paper incorporates the real histopathological images into the model training process by proposing a self-supervised consistency regularization. Specifically, the contributions of this paper are listed as below.

\begin{itemize}
    \item A novel weakly-supervised tissue segmentation framework, named HisynSeg, is proposed. With an image-mixing synthesis approach, the framework converts the weakly-supervised task into a fully-supervised one, thus evading the existing issues in CAM-based WSSS methods.
    
    \item Two image-mixing synthesis strategies are proposed based on Mosaic transformation and Bézier mask generation. In addition, a synthesized image filtering module is developed to further ensure the authenticity of the synthesized images.
    
    \item To avoid the occasional artifacts in the synthesized images, we design a self-supervised consistency regularization that allows real images without pixel-level masks to participate in the model training process, significantly improving the performance of the proposed framework.
    
    \item Experiments on three WSSS datasets with histopathological images validate that the proposed framework has a large performance improvement compared to the state-of-the-art (SOTA) methods.
    
\end{itemize}

\section{Related Works}
\label{sec:related}
\subsection{Weakly-Supervised Semantic Segmentation}
\subsubsection{WSSS for Natural Images}
Due to the time-consuming and laborious process of obtaining pixel-level segmentation masks, as well as the risk of inaccurate annotation, many researchers have attempted to achieve semantic segmentation using weak labels. Among all types of weak labels, image-level classification labels have received extensive attention due to their ease of access. Most existing studies first train a classification network and then generate pseudo-masks for the training set based on the CAM \cite{zhou2016learning} to enable segmentation model training. These studies mainly focus on designing novel CAM generation strategies or training constraints to amend the existing issues in CAM. For example, to solve the under-activation problem of CAM, SC-CAM \cite{chang2020weakly} divides each classification category into subclasses by clustering. Ultimately, the CAMs of different subclasses are fused, improving the accuracy of segmentation boundaries. Considering the equivariant property of the segmentation task, Wang et al. \cite{wang2020self} design a self-supervised equivariant attention mechanism (SEAM) to alleviate the under-activation and over-activation problems in CAM. However, the above studies are all designed for natural images. Compared to natural images, histopathological images have more homogeneous content and more uniform colors, which often makes the performance of methods designed for natural images unsatisfactory when directly applied to histopathological images.

\subsubsection{WSSS for Histopathological Images}
In WSSS for histopathological images, most studies are also based on CAM. HistoSegNet \cite{chan2019histosegnet} is one of the earliest works focusing on weakly-supervised histopathology tissue segmentation. HistoSegNet firstly infers pseudo-masks with GradCAM \cite{8237336} and then designs a series of post-processing strategies to refine the pseudo-masks. However, the manually-defined post-processing methods are data-sensitive, thereby limiting the generalization ability of the model. Han et al. \cite{han2022multi} propose a novel WSSS model named WSSS-Tissue for histopathological images with a progressive dropout attention strategy, forcing the model to focus on the boundary regions of the tissues. To tackle the low-resolution problem of CAMs, Zhong et al. \cite{zhonghamil} propose a WSSS framework named HAMIL. Li et al. \cite{li2022online} propose an online easy example mining mechanism named OEEM, where the segmentation loss is weighted by the confidence of the predicted probability map to avoid incorrect predictions in the pseudo-masks affecting the segmentation performance. 
Zhong et al. \cite{zhong2024semi} introduce a CDMA+ framework by adopting a multi-task learning approach, where weakly-supervised segmentation is considered as an auxiliary task while semi-supervised segmentation is taken as the primary task. Specifically, CDMA+ employs CAM to constrain the learning process of the segmentation probability maps, which is similar to the proposed consistency regularization in this paper. However, CDMA+ utilizes upsampling to interpolate the CAMs to the same shape as the probability maps to compute consistency. In contrast, we use downsampling to reshape the probability maps instead of the CAMs. Considering that downsampling usually more accurately preserve the contour of the image content than interpolation, the consistency constraint proposed in this work may be more reasonable. Besides, all the abovementioned studies are still improvements to CAM and do not consider the characteristics like more uniform colors of histopathological images. 
In contrast, this paper fully uses the homogeneity property of the histopathological images to generate synthesized images with pixel-level masks, which provides new clues to WSSS for histopathological images.

In addition to CAM-based methods, some studies utilize multiple-instance learning (MIL) for weakly-supervised tissue segmentation. For example, Xu et al. \cite{xu2019camel} propose a CAMEL method, which takes the histopathological images as bags and the cropped small patches in the images as instances. Specifically, CAMEL first proposes an instance pseudo-labeling module by developing a cMIL strategy to obtain the pseudo-labels for the patches. Next, fine-grained pseudo-masks are obtained to train a segmentation model by assigning the pseudo-labels to all pixels in the patch. By taking histopathological images as bags and the pixels in the images as instances, Jia et al. \cite{jia2017constrained} propose a framework named DWS-MIL for cancerous and non-cancerous tissue segmentation. However, the MIL-based WSSS methods only consider two tissue categories. On the contrary, in clinical, clinicians usually need to analyze multiple tissue types in the TME during cancer treatment, which limits the application of these MIL-based methods.

\subsection{Image Synthesis in Computer Vision}
Image synthesis is one of the most important tasks in computer vision. Since Goodfellow et al. \cite{goodfellowgenerative} proposed the generative adversarial network (GAN), researchers have conducted a series of explorations in image synthesis based on GAN. For example, the Pix2Pix \cite{isola2017image} is first proposed to accomplish the image-to-image synthesis task. The CycleGAN \cite{zhu2017unpaired} is proposed to solve the problem that paired data required by Pix2Pix is difficult to obtain. In medical image synthesis, Guan et al. \cite{guanunsupervised} design a GramGAN framework to synthesize histopathological images with different stainings from the hematoxylin-eosin (H\&E) stained images. In recent years, diffusion models have attracted extensive attention due to their advantages of a more stable training process and more diverse generation styles \cite{ho2020denoising, choi2021ilvr, rombach2022high}. In the medical field, Moghadam et al. \cite{moghadam2023morphology} achieve histopathological image synthesis based on a diffusion probability model using genotype labels as conditions. Oh et al. \cite{ohdiffmix} propose a method called DiffMix, which uses pixel-level cell segmentation masks as conditions and synthesizes histopathological images corresponding to these masks. 
However, generative model-based methods cannot be directly applied to WSSS tasks, because they either only synthesize histopathological images without segmentation masks or require known segmentation masks as conditions to generate the corresponding images.
\begin{figure*}[!t]
    \centering
    \includegraphics[width=0.85\linewidth]{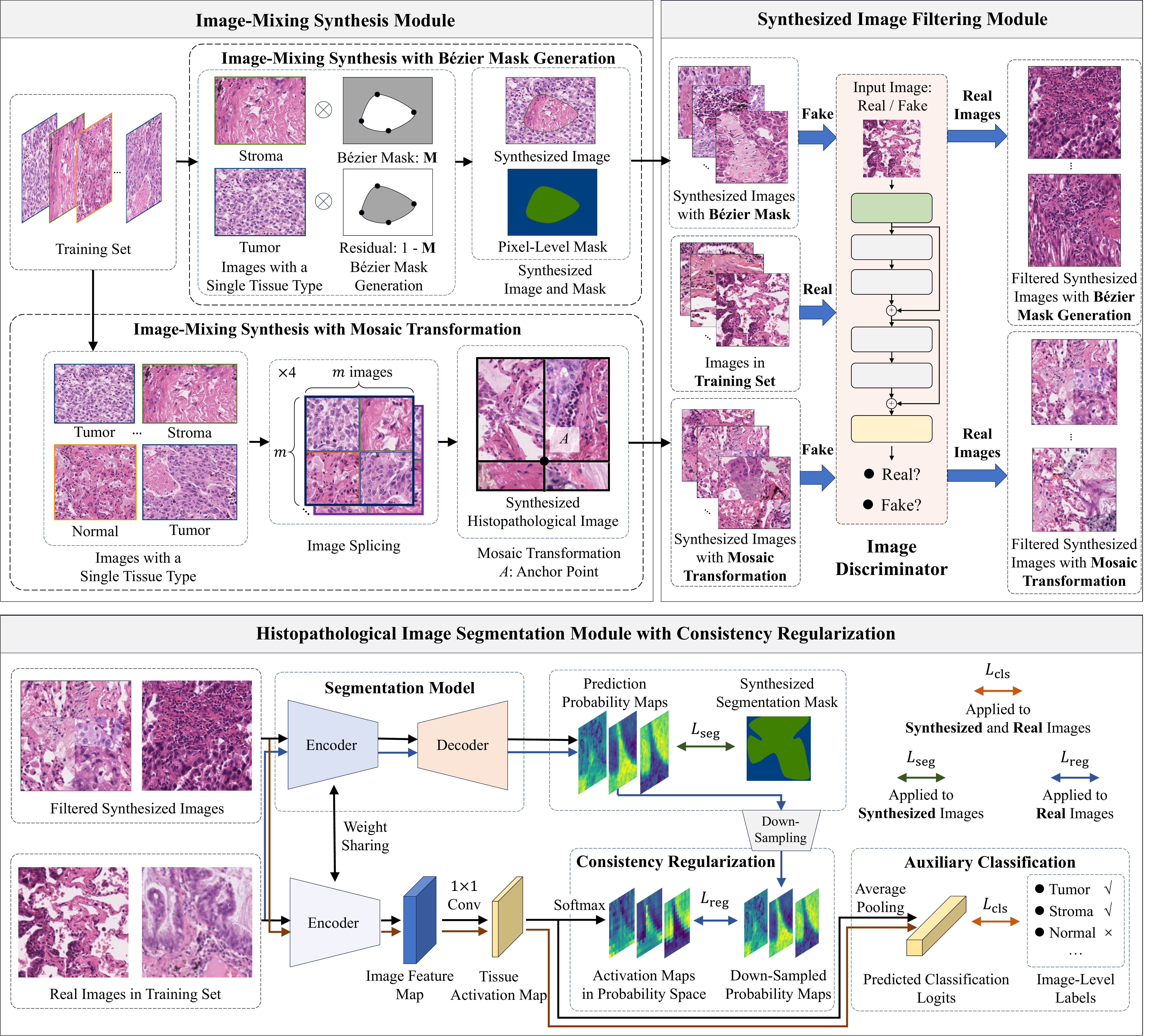}
    \caption{An overview of the HisynSeg framework. The framework is composed of three modules. In the image-mixing synthesis module, synthesized images and masks are generated by two proposed strategies, namely Mosaic transformation and Bézier mask generation. Next, the synthesized image filtering module selects authentic images from the synthesized images. Finally, a segmentation model is trained in the histopathological image segmentation module. To avoid the occasional artifacts in the synthesized images affecting the performance of the segmentation model, real images are fed into the model, which is trained using a proposed consistency regularization under a self-supervised scheme.}
    \label{fig:2}
\end{figure*}

In addition, there are also some works that conduct image synthesis without generative models. For example, CutMix \cite{yun2019cutmix} mixes two different images by cropping a rectangular region of one image and pasting it onto another. 
However, there will be a series of problems and limitations if directly applying CutMix to the field of weakly-supervised histopathological image segmentation. For instance, the rectangular cropping operation in CutMix causes the seams to always be parallel to the axes. Furthermore, the four orthogonal corners in the cropped region cannot reflect the smooth, continuous boundaries usually observed in the real TME. Compared to CutMix, the two synthesis strategies proposed in this paper can solve the above problems, leading to better performance.

\section{Method}
\label{sec:Method}
To avoid the issues in CAMs, we propose a HisynSeg framework empowered by an image-mixing synthesis approach. An overview of HisynSeg is shown in Fig. \ref{fig:2}. The framework consists of three modules, i.e., an image-mixing synthesis module, a synthesized image filtering module, and a histopathological image segmentation module. In the image-mixing synthesis module, we propose two synthesis strategies: Mosaic transformation and Bézier mask generation. The role of the synthesized image filtering module is to pick out authentic images from all the synthesized images. The histopathological image segmentation module is built to train a segmentation model in a fully-supervised manner with the synthesized images and corresponding pixel-level masks. Besides, to improve the robustness of the model to the occasional artifacts in the synthesized images, the real images in the training set are also fed into the segmentation model. Considering that real images do not have segmentation masks, this paper further proposes a self-supervised consistency regularization. The details of each module are explained in the following subsections. 

\subsection{Image-Mixing Synthesis Module}
Unlike natural images, whose backgrounds often have semantic information (e.g., sky, grass, and water), the backgrounds of histopathological images are often white regions. Therefore, we can obtain accurate semantic segmentation masks for histopathological images with a single tissue type. However, directly using these images to train a segmentation model is unreasonable because multiple tissue types exist in real TMEs. Considering the accessibility of the pixel-level masks for the images with a single tissue type, it is appropriate to leverage them to synthesize histopathological images with multiple tissue types. 
Therefore, to make full use of the sufficient semantic information contained in histopathological images with a single tissue type, we propose two image-mixing synthesis strategies, namely Mosaic transformation and Bézier mask generation. 

\subsubsection{Image-Mixing Synthesis with Mosaic Transformation}
The Mosaic transformation \cite{bochkovskiy2020yolov4} commonly utilizes four images $\{I_i\}_{i=1}^4$ to generate one synthesized image. For a synthesized image $I^\prime_\text{M}$ shaped $H\times W$, the Mosaic transformation first arbitrarily selects an anchor point $A = (H_A, W_A)$, where $\alpha H < H_A < \beta H$ and $\alpha W < W_A < \beta W $ ($0 < \alpha < \beta < 1$. We set $\alpha=0.2$ and $\beta=0.8$ in the experiments). Then, the four parts divided by the anchor point in the synthesized image are filled by the pixels in $\{I_i\}_{i=1}^4$, respectively. More specifically, the four images $\{I_i\}_{i=1}^4$ are firstly randomly cropped into four intermediate images $\{I_i^{\prime\prime}\}_{i=1}^4$, whose shapes are $H_A\times W_A$, $(H-H_A) \times W_A$, $H_A\times(W-W_A)$, and $(H-H_A)\times (W-W_A)$. Then, the four intermediate images are placed on the four corners of the synthesized image, i.e.,
\begin{equation}
    I^\prime_\text{M} = \begin{bmatrix}
    I_1^{\prime\prime} & I_3^{\prime\prime} \\
    I_2^{\prime\prime} & I_4^{\prime\prime}
    \end{bmatrix}.
\end{equation}

In this paper, each image used for Mosaic transformation, i.e., $I_i$, is generated by splicing histopathological images with a single tissue type. Firstly, a total number of $m^2$ images with a single tissue type are arbitrarily selected from the dataset ($m=2$ in experiments). Next, each selected image is randomly cropped to a shape of $\frac{H}{m}\times \frac{W}{m}$. Then, a gridded image can be obtained by splicing the $m^2$ images in a raster scanning order. Finally, the gridded images are utilized for Mosaic transformation after random rotation and scaling.

\subsubsection{Image-Mixing Synthesis with Bézier Mask Generation}
\begin{figure}
    \centering
    \includegraphics[width=0.95\linewidth]{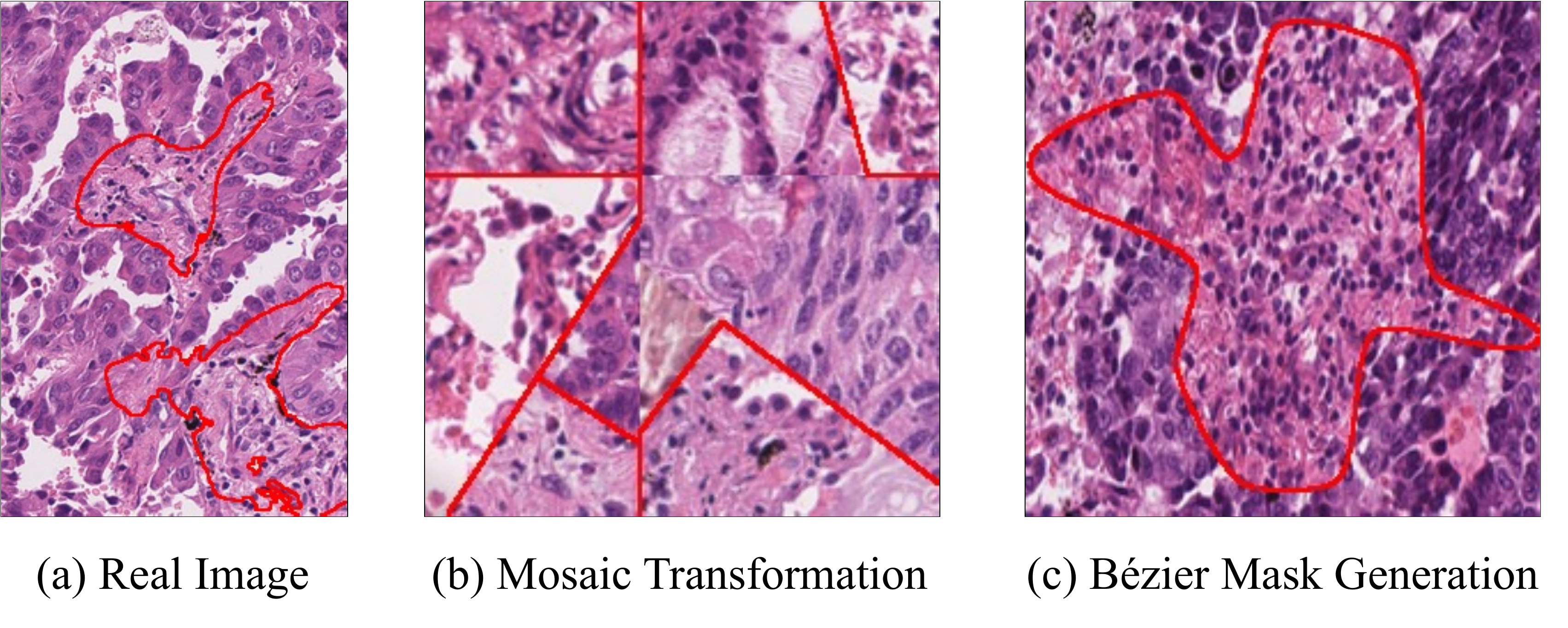}
    \caption{The segmentation boundaries between different types of tissues for (a) a real image, (b) a synthesized image by Mosaic transformation, and (c) a synthesized image by Bézier mask generation. The red lines/curves in the images represent the segmentation boundaries.}
    \label{fig:r1-3}
\end{figure}

Although the image-mixing synthesis strategy based on Mosaic transformation can provide histopathological images with multiple tissue types, introducing abundant tissue heterogeneity and semantic information about inter- and intra-tissue interactions into the synthesized images, the splicing operation in Mosaic transformation makes the segmentation boundary between different types of tissues a straight line, as shown in Fig. \ref{fig:r1-3}. However, the boundaries between different types of tissues in real histopathological images are mostly smooth curves. To this end, this paper further designs an image-mixing synthesis strategy by generating masks with smooth Bézier curves \cite{elberinterpolation}. Benefiting from the smoothness of the Bézier curves, the synthesized images have similar appearances to the real images, thereby improving the segmentation performance.

Specifically, we achieve image-mixing synthesis by constructing multiple consecutive Bézier curves to form a closed shape, which is used as a Bézier mask. Formally, an $n$-order Bézier curve can be uniquely determined by $n+1$ control points, denoted as $\mathbf{P}_0, \mathbf{P}_1, \cdots, \mathbf{P}_n$. The curve equation $\mathbf{C}(t)$ with parameter $t$ can be calculated by
\begin{equation}
    \mathbf{C}(t) = \sum_{k=0}^n \binom{n}{k}t^k(1-t)^{n-k}\mathbf{P}_k, \quad t\in [0, 1],
\end{equation}
where $\mathbf{P}_k = [x_k, y_k]^\top$ is the coordinate of a control point, $\binom{n}{k} = \frac{n!}{k!(n-k)!}$ is the bionormial coefficient, and $0^0$ is defined as $1$ \cite{elberinterpolation}.

Next, multiple Bézier curves are connected end-to-end to construct the Bézier mask. Firstly, $N$ points are randomly taken in a unit square following the anti-clockwise direction. Then, two consecutive points $\mathbf{P}_i$ and $\mathbf{P}_{(i+1) \bmod N}$ are taken in turn as the start and terminal control points of a $3$-order Bézier curve $\mathbf{C}_i(t)$ ($i=0,\cdots, N-1$). Since the $3$-order Bézier curve needs to be determined by four control points, the two other control points are generated between the two selected points to guarantee the smoothness between two successive Bézier curves. In other words, the first derivative at the endpoint of the former Bézier curve should be equal to the first derivative at the start point of the latter Bézier curve, i.e., 
\begin{equation}
    \frac{\mathrm{d}\mathbf{C}_i(t)}{\mathrm{d}t}\bigg|_{t=1}  = \frac{\mathrm{d} \mathbf{C}_{(i+1)\bmod N}(t)}{\mathrm{d}t}\bigg|_{t=0},\, i=0,\cdots,N-1.
\end{equation}

Finally, by scaling the coordinates of the curves to the image size, a smooth Bézier mask $\mathbf{M}$ composed of $N$ Bézier curves can be generated. By assigning $1$ to the inner part and $0$ to the outer part of the mask, a synthesized image can be generated with two histopathological images with a single tissue type by
\begin{equation}
    I^\prime_\text{B} = \mathbf{M} \otimes I_1 + (1-\mathbf{M}) \otimes I_2,
\end{equation}
where $\otimes$ means element-wise multiplication, and $I_1$ and $I_2$ represent the two selected histopathological images.

Since the synthesized images are made up of images with a single tissue type, for each synthesized image, we can obtain the pixel-level segmentation mask, which is utilized as the ground truth for training the segmentation model.

\subsection{Synthesized Image Filtering Module}
In the synthesized images, artifacts usually exist due to variations in staining concentration and brightness. Besides, histopathological images with different subtypes, grades, or other clinical characteristics may be composed in a synthesized image, making the TME features in the synthesized image differ from those in the real histopathological images. Therefore, if we directly utilize the synthesized images as training samples for the segmentation model, there is a large probability that the model overfits these artifacts, leading to a decline in the segmentation performance. 
\begin{figure}
    \centering
    \includegraphics[width=\linewidth]{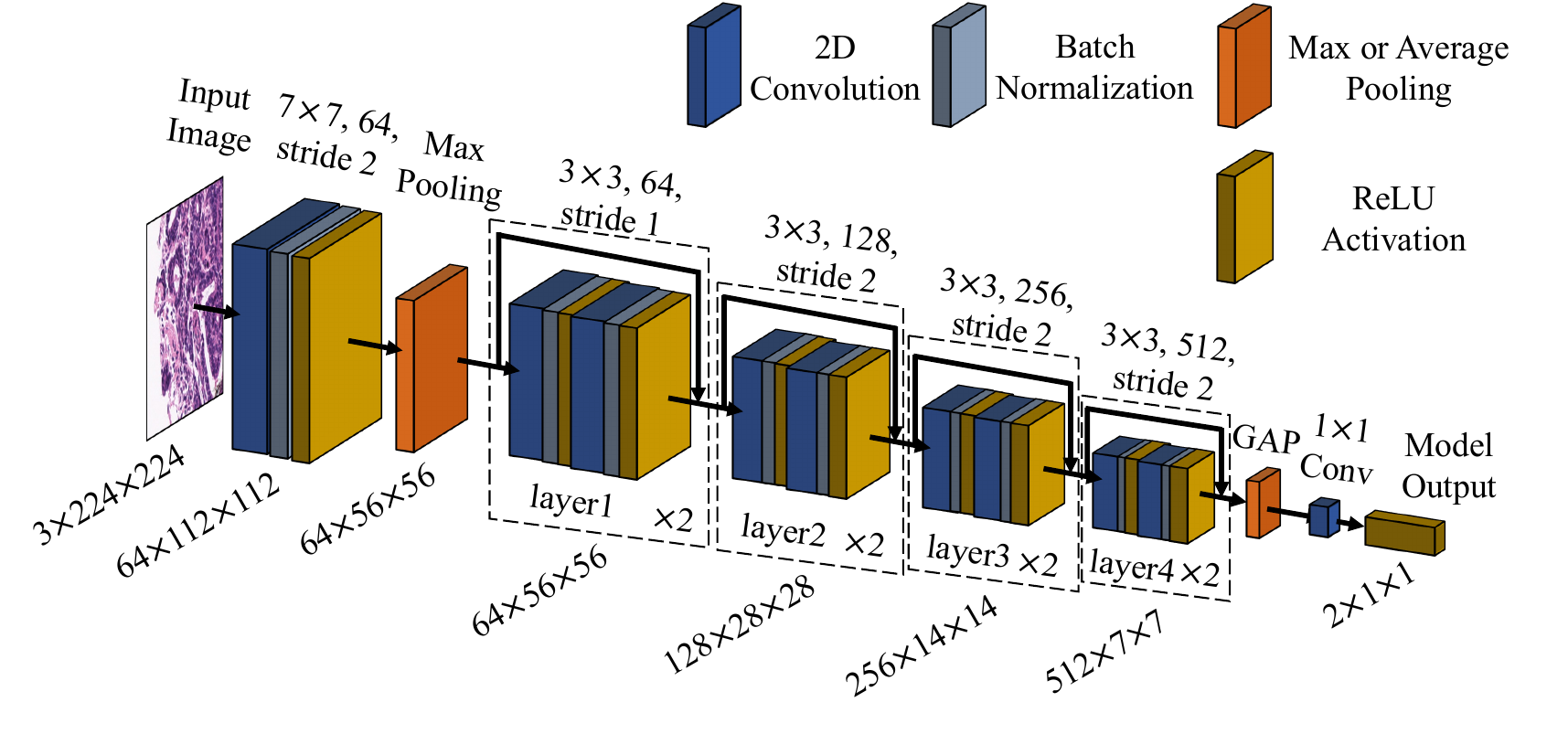}
    \caption{The architecture of the discriminator, which is based on the backbone of ResNet-18. The last fully connected layer (i.e., $1\times 1$ convolution) of our discriminator is modified to output a 2D vector for real and fake image classification. The kernel shapes and strides of the first convolution in each layer are listed above each block. The output shapes are noted below each block. GAP: Global Average Pooling.}
    \label{fig:r1-4}
\end{figure}

For this reason, this paper designs a synthesized image filtering module to provide synthesized images that are the most similar to the real images. The synthesized image filtering module has a similar function as an image discriminator, which discriminates whether an input image is from the WSSS dataset (real) or the synthesized one (fake). Specifically, a ResNet-18 \cite{he2016deep} model is utilized as the discriminator, whose structure is shown in Fig. \ref{fig:r1-4}.

To train the discriminator, we first employ the synthesized images as negative samples with label $0$ and utilize images from the WSSS dataset as positive samples with label $1$. Specifically, to maintain consistency with the fact that the synthesized images contain multiple tissue types, we only select images with two or more tissue types in the WSSS dataset. 
The cross-entropy loss is used to optimize the discriminator, which outputs a probability indicating whether an image is fake or real. In the inference stage, for an arbitary synthesized image, if the discriminator gives a high possibility ($>0.5$), we save it as an authentic image.
Otherwise, the synthesized image is abandoned. With the proposed approach, the consistency of the synthesized images with the real ones can be ensured, and artifacts in the synthesized images can be reduced by discarding synthesized images with low probabilities of being real. It is worth noting that for better discrimination, we separately train two discriminators for Mosaic transformation and Bézier mask generation, respectively.

\subsection{Segmentation with Consistency Regularization}
After obtaining the authentic synthesized images from the image filtering module, we can train a tissue segmentation model under a fully-supervised scheme. Denote a synthesized image as $I_{\text{syn}}$, whose segmentation mask is $M_{\text{syn}}$, the predicted segmentation probability map can be obtained by
\begin{equation}
    \hat{M}_{\text{syn}} = \mathcal{D}(\mathcal{E}(I_{\text{syn}})),
\end{equation}
where $\mathcal{D}$ and $\mathcal{E}$ stand for the decoder and the encoder of the segmentation model, respectively. Then, the segmentation probability map can be supervised via the segmentation loss, which is defined by
\begin{equation}
    L_\text{seg} = \text{Dice}(\hat{M}_{\text{syn}}, M_{\text{syn}}),
\end{equation}
where $\text{Dice}(x, y) = 1 - \frac{ 2\lvert x \cap  y  \rvert }{\lvert x \rvert + \lvert y  \rvert}$ represents the Dice loss.

Although the authenticity of the synthesized images has greatly improved with the filtering module, it is still difficult to guarantee that there are no artifacts in the filtered synthesized images. In contrast, real images in the WSSS dataset do not contain any synthesis artifacts and have true TME. Therefore, leveraging the real images for model training can further prevent the model from overfitting the occasional artifacts while facilitating the model to capture the heterogeneity of the TME. 
Inspired by this, we utilize the real histopathological images in the WSSS dataset to supervise the segmentation model. Denote a real image as $I_\text{real}$. By feeding it into the segmentation network, the predicted segmentation probability map can be yielded by 
\begin{equation}
    \hat{M}_{\text{real}} = \mathcal{D}(\mathcal{E}(I_{\text{real}})).
\end{equation}

However, the real images only have image-level classification labels. Therefore, no pixel-level segmentation masks can be utilized to supervise the predicted probability map $\hat{M}_{\text{real}}$. To solve this problem, we adopt the idea of CAM, where the image features extracted by a classification network are utilized to locate the semantic information in an image. Specifically, an auxiliary classification task is built to obtain a tissue activation map to supervise the probability map $\hat{M}_{\text{real}}$. 

We take the encoder $\mathcal{E}$ as the backbone of a classification network (i.e., sharing the same network structure and weights with $\mathcal{E}$) and add a specially designed classification head after the last feature extraction layer of the encoder. Specifically, we build a $1\times 1$ convolution layer to project the feature map to an output category space, i.e., 
\begin{equation}
    \mathbf{F}_c = \text{Conv}_{1\times 1}(\mathbf{F}).
\end{equation}
Here, the feature map extracted by the last layer of the encoder is denoted as $\mathbf{F} \in \mathbb{R}^{H^\prime\times W^\prime \times D}$, where $H^\prime$, $W^\prime$, and $D$ stand for the height, width, and channel number of the feature map, respectively. The output $\mathbf{F}_c\in \mathbb{R}^{H^\prime \times W^\prime \times C}$ has the same number of channels as the segmentation map (i.e., number of tissue categories). Therefore, it can be viewed as a tissue activation map, which should share a consistency with the segmentation probability map $\hat{M}_{\text{real}}$. To this end, a consistency regularization is proposed to enable the supervision of $\hat{M}_{\text{real}}$ with the L1 loss by
\begin{equation}
    L_\text{reg} = \lVert \sigma(\mathbf{F}_c) - R(\hat{M}_{\text{real}}) \rVert_1,
\end{equation}
where $\sigma$ and $R$ represent the softmax function and the down-sampling operation, respectively. Here, considering the distributions of $\hat{M}_{\text{real}}$ and $\mathbf{F}_c$ are different, we first project $\mathbf{F}_c$ into a probability space using the softmax activation and then guarantee the pixel-wise consistency with the L1 norm by down-sampling $\hat{M}_\text{real}$ to make the tensors' shapes compatible. Because the synthesized images have more accurate masks than the activation map, $L_\text{reg}$ is only calculated for real images without pixel-level masks.

Next, to complete the auxiliary classification task, an average pooling operation is applied to the activation map, and the classification logits of the image can be obtained by
\begin{equation}
    z = P (\mathbf{F}_c),
\end{equation}
where $P$ is the average pooling operation.

Finally, the multi-label soft margin loss is adopted to supervise the classification task as
\begin{equation}
    L_\text{cls} = -\frac{1}{C} \sum_{i=1}^{C} y_i \log \frac{1}{1+e^{-z_i}} + (1-y_i) \log \frac{e^{-z_i}}{1+e^{-z_i}},
\end{equation}
where $y_i \in \{0, 1\}$ is the ground-truth label of category $i$, and $z_i$ is the predicted logits of category $i$ by the classification head. Both real and synthesized images take part in the calculation of the classification loss, as both image-level labels can be obtained from the classification labels or the synthesized masks.

To conclude, the total loss for training the segmentation model is defined as the summation of segmentation loss, consistency regularization, and classification loss, i.e.,
\begin{equation}
    L = L_\text{seg}+ L_\text{reg} + L_\text{cls}.
\end{equation}

\subsection{Iterative Training Strategy}
After training the segmentation model based on $L$, the model can be used to generate pseudo-masks for the real images in the WSSS dataset, which can be utilized to supervise the segmentation probability maps of the real images, enabling an iterative training strategy for the segmentation model. During the iterative training process, since the training images already have pseudo-masks, we no longer use the consistency regularization $L_\text{reg}$ but directly compute the segmentation loss $L_\text{seg}$ and the classification loss $L_\text{cls}$ for both real and synthesized images.

\section{Experiments and Results}
\label{sec:experiments}
\subsection{Experiment Dataset}
This paper uses three weakly-supervised tissue semantic segmentation datasets for experimental evaluation. The details are described as follows.

\paragraph{WSSS4LUAD \cite{hanwsssluad}} The dataset is derived from two cohorts from the Cancer Genome Atlas and Guangdong Provincial People's Hospital, respectively. The training set contains 10,091 histopathological images with sizes ranging from $(200\sim500) \times (200\sim500)$ pixels. For each image in the training set, the existence of three tissue types, i.e., tumor (TUM), stroma (STR), and normal (NOM) tissue, is annotated in a multi-label scheme. The validation set consists of 31 small images ranging from $(200\sim500) \times (200\sim500)$ pixels and 9 large images ranging from $(1500\sim5000) \times (1500\sim5000)$ pixels. The test set contains 66 small and 14 large images whose shape range is the same as the validation set. 

The annotation process of WSSS4LUAD was completed by eight junior, two senior, and one expert pathologist. Specifically, the junior pathologists were responsible for annotation and labeling, while the senior and expert pathologists were required to check and verify the results. Deep learning models were utilized to accelerate the labeling and annotation process. A ResNet-38 model was first trained with around 500 patches with pathologist-annotated image-level labels. The trained ResNet-38 model was then utilized to infer pseudo-labels for other patches. Patches with low prediction confidence were reviewed and corrected by the senior pathologists, and the junior pathologists checked those with high confidence. For pixel-level annotation, the deep learning framework WSSS-Tissue \cite{han2022multi} was utilized to generate initial pseudo-masks for each image. Then, PhotoShop was used to refine the pseudo-masks. A label review board confirmed all the image-level labels and pixel-level masks.

\paragraph{BCSS-WSSS} This dataset is a WSSS version of the BCSS dataset \cite{amgad2019structured}, which was created by Han et al. \cite{han2022multi}. The original BCSS dataset consists of 151 WSIs, each of which was annotated with a region of interest (RoI). The RoIs were selected and annotated with pixel-level masks by a study coordinator and a doctor. A senior pathologist was responsible for checking and approving the results. The original BCSS dataset was annotated with 21 different tissue types. The BCSS-WSSS dataset combined these 21 tissue types and background pixels into 5 classes, namely tumor (TUM), stroma (STR), lymphocytic infiltrate (LYM), necrosis (NEC), and background. Finally, by randomly cropping $224\times 224$ patches from the RoIs, Han et al. \cite{han2022multi} built the training, validation, and test sets which consist of 23,422, 3,418, and 4,986 histopathological images, respectively. The image-level labels were inferred from the pixel-level masks.

\paragraph{LUAD-HistoSeg \cite{han2022multi}} This dataset is derived from the H\&E stained WSIs of lung adenocarcinoma patients from Guangdong Provincial People's Hospital. $224\times 224$-shaped images are cropped from the WSIs, making the LUAD-HistoSeg dataset consist of a training set of 16,678, a validation set of 300, and a test set of 307 images. Four tissue types, tumor epithelial (TE), necrosis (NEC), lymphocyte (LYM), and tumor-associated stroma (TAS), are labeled in the dataset. 
The dataset was annotated by five junior clinicians and three professional pathologists. The junior clinicians were responsible for annotation with a two-stage process. In the first stage, the junior clinicians utilized the Labelme software to outline rough segmentation masks. Next, PhotoShop was used to refine the masks. After labeling and annotating all images, the three professional pathologists were asked to check the labels and annotations. Images with poor quality (e.g., blurred, dirty, and unevenly stained) or suffering from inconsistency among the pathologists' annotations were discarded.

Since HisynSeg needs images with a single tissue type to synthesize images, we list the number of these images in Table \ref{tab:R1.2}. Although the number of images with a single tissue type is not abundant for some datasets, e.g., LYM images in the LUAD-HistoSeg dataset, the following experimental results validate that the proposed HisynSeg can still have competitive performance compared with the SOTA frameworks.
\begin{table}[!t]
\centering
\caption{The number of images with a single tissue type in the three datasets.}
\label{tab:R1.2}
\begin{tabular}{@{}ccc@{}}
\toprule
Dataset                                             & \multicolumn{1}{c}{Tissue Type}                    & Image Number \\ \midrule
\multicolumn{1}{c}{\multirow{3}{*}{WSSS4LUAD}}     & \multicolumn{1}{c}{Tumor (TUM)}                   & 1181         \\
\multicolumn{1}{c}{}                               & \multicolumn{1}{c}{Stroma (STR)}                  & 1680         \\
\multicolumn{1}{c}{}                               & \multicolumn{1}{c}{Normal (NOM)}                  & 1832         \\ \midrule
\multicolumn{1}{c}{\multirow{4}{*}{BCSS-WSSS}}     & \multicolumn{1}{c}{Tumor (TUM)}                   & 4738         \\
\multicolumn{1}{c}{}                               & \multicolumn{1}{c}{Stroma (STR)}                  & 2903         \\
\multicolumn{1}{c}{}                               & \multicolumn{1}{c}{Lymphocytic Infiltrate (LYM)}  & 679          \\
\multicolumn{1}{c}{}                               & \multicolumn{1}{c}{Necrosis (NEC)}                & 1058         \\ \midrule
\multicolumn{1}{c}{\multirow{4}{*}{LUAD-HistoSeg}} & \multicolumn{1}{c}{Tumor Epithelial (TE)}         & 1574         \\
\multicolumn{1}{c}{}                               & \multicolumn{1}{c}{Necrosis (NEC)}                & 787          \\
\multicolumn{1}{c}{}                               & \multicolumn{1}{c}{Lymphocyte (LYM)}              & 42           \\
\multicolumn{1}{c}{}                               & \multicolumn{1}{c}{Tumor-Associated Stroma (TAS)} & 2192         \\ \bottomrule
\end{tabular}
\end{table}

\subsection{Evaluation Metrics}
The experimental evaluation utilizes two common segmentation metrics, mean intersection over union (mIoU) and frequency-weighted IoU (fwIoU), with mIoU being the primary metric since it best reflects the overall performance. Meanwhile, the IoU for each tissue category is also reported. It should be noted that because the white background regions in histopathological images can be easily obtained by color thresholding and can be viewed as known, we do not consider these background pixels when calculating the evaluation metrics as in Ref. \cite{han2022multi}. 

\subsection{Experimental Settings and Implementation Details}
All experiments are conducted on a Ubuntu 18.04 LTS server with a single Nvidia RTX 3090 GPU. All codes are implemented using Pytorch 1.12.1 and Pytorch Lightning 1.7.1. For the WSSS4LUAD dataset and the LUAD-HistoSeg dataset, we separately synthesize 3,600 real images with the Mosaic transformation strategy or the Bézier mask generation strategy. For the BCSS-WSSS dataset, 7,200 real images are synthesized with each strategy. Since the WSSS4LUAD dataset focuses more on the segmentation of tumors and stroma, the foreground and background images used in the Bézier mask generation strategy are only selected from either tumor or stroma images. For other datasets, the images utilized for synthesizing are randomly selected from images with a single tissue type. For Mosaic transformation, the four gridded images are preprocessed by random flipping, shifting, scaling, rotation, and cropping. We set $H=W=224$ in Mosaic transformation to ensure the synthesized images are $224\times 224$. In contrast, there is no pre-processing for Bézier mask generation, except in the WSSS4LUAD dataset, where we first use bilinear interpolation to reshape the images into $224\times 224$ to solve the problem of different image shapes. For the synthesized image filtering module, a ResNet-18 \cite{he2016deep} is trained from scratch for each synthesized dataset with an Adam \cite{kingmaadam} optimizer and a 0.001 learning rate. For the BCSS-WSSS dataset, we train the discriminator for three epochs on the synthesized dataset with Mosaic transformation. For other synthesized datasets, we train the discriminator for five epochs. The reason for utilizing a small epoch number mainly comes from the fact that abundant number of real images (over thousands for each dataset) and synthesized images can be obtained to train the discriminator. Therefore, even with a small epoch number, the discriminator can still be trained well to learn the differences between the real and synthesized images, thereby improving the performance of the segmentation network.

The segmentation network is implemented by DeepLabV3+ \cite{chen2018encoder} with EfficientNet-b6 \cite{tan2019EfficientNet} as the backbone. The network is trained for 60 epochs over the WSSS4LUAD dataset and 25 epochs over other datasets. The optimizer is AdamW \cite{loshchilov2018decoupled} with a learning rate of 0.0001 and a weight decay of 0.05. The model with the best validation mIoU is selected for testing. During training, random resizing, cropping, flipping, shifting, scaling, rotation, and optical distortion are utilized as pre-processing to augment the training images. For the WSSS4LUAD dataset, we resize all training images to $224\times 224$ by bilinear interpolation. Besides, we utilize a sliding window strategy with a window size of $224\times 224$ in inference to handle the large-size validation and test images. Specifically, no overlapping is taken in validation. While in testing, we use multiscale images with 50\% overlapping. The sliding window strategy is also utilized in all the competing methods for fairness. In addition, we adopt a test-time augmentation strategy with flips and 0, 90, 180, and 270-degree rotations. The final results are fused by averaging. We adopt the iterative training strategy over the BCSS-WSSS and the LUAD-HistoSeg dataset and retrain the model for one time. We do not utilize the iterative training strategy over the WSSS4LUAD dataset since we empirically find no performance improvement. All baselines are compared strictly following the published papers or open-sourced codes.

\subsection{Statistical Analysis}
In order to alleviate the impact of randomness on performance evaluation, we use five seeds for each experiment to train five models with the same training, validation, and test set, and report the average value with standard deviation for each metric in the form of $\text{Mean}_{\text{std}}$, where $\text{Mean}$ and $\text{std}$ represent the average of the metric and the standard deviation, respectively.
In this paper, a two-tailed permutation test is utilized to calculate the p-value of the difference between the performance of our proposed HisynSeg and the competing methods. 
It is worth noting that the permutation test has been widely used to calculate p-values of different performance metrics in computational pathology \cite{chen2024towards, lu2024visual}. 
Unless otherwise stated, we use * in each reported metric of a competing method to represent that the metric of the competing method is significantly different from ours (the last line of a table) with a p-value less than 0.05.

\subsection{Comparative Experiments}
\subsubsection{Quantitative Results}
\begin{table*}[!t]
\setlength{\tabcolsep}{1pt}
\centering
\caption{Performance comparison over WSSS4LUAD and BCSS-WSSS datasets. We bold the best and underline the second-best results.}
\label{tab:2}
\begin{tabular}{@{}cccccc|cccccc@{}}
\toprule
\multirow{2}{*}{Method}                  & \multicolumn{5}{c|}{WSSS4LUAD}                                                                                                                               & \multicolumn{6}{c}{BCSS-WSSS}                                                                                                                                                              \\ \cmidrule(l){2-12} 
                                         & \multicolumn{1}{c}{TUM (\%)} & \multicolumn{1}{c}{STR (\%)} & \multicolumn{1}{c}{NOM (\%)} & \multicolumn{1}{c}{mIoU (\%)} & \multicolumn{1}{c|}{fwIoU (\%)} & \multicolumn{1}{c}{TUM (\%)} & \multicolumn{1}{c}{STR (\%)} & \multicolumn{1}{c}{LYM (\%)} & \multicolumn{1}{c}{NEC (\%)} & \multicolumn{1}{c}{mIoU (\%)} & \multicolumn{1}{c}{fwIoU (\%)} \\ \midrule 
GradCAM   \cite{8237336}                 & $67.12_{7.15}^\ast$          & $52.01_{4.68}^\ast$          & $15.58_{20.45}^\ast$         & $44.90_{10.04}^\ast$          & $59.42_{6.17}^\ast$             & $24.97_{13.45}^\ast$         & $21.42_{10.92}^\ast$         & $11.87_{5.84}^\ast$          & $2.64_{1.35}^\ast$           & $15.22_{3.41}^\ast$           & $21.23_{7.32}^\ast$            \\
HistoSegNet   \cite{chan2019histosegnet} & $42.02_{11.68}^\ast$         & $42.96_{3.10}^\ast$          & $62.07_{3.05}^\ast$          & $49.02_{2.67}^\ast$           & $43.00_{5.57}^\ast$             & $55.23_{2.27}^\ast$          & $59.53_{1.06}^\ast$          & $30.14_{6.44}^\ast$          & $58.14_{3.52}^\ast$          & $50.76_{2.06}^\ast$           & $54.71_{1.24}^\ast$            \\
SEAM   \cite{wang2020self}               & $68.22_{10.75}^\ast$         & $58.86_{13.82}^\ast$         & $64.98_{6.14}^\ast$          & $64.02_{5.73}^\ast$           & $64.30_{9.93}^\ast$             & $78.53_{2.51}^\ast$          & $70.92_{4.53}^\ast$          & $56.24_{5.29}$               & $61.82_{7.41}$               & $66.88_{4.60}^\ast$           & $72.36_{3.69}^\ast$            \\
SC-CAM   \cite{chang2020weakly}          & $\uline{80.03_{0.65}^\ast}$  & $\uline{74.13_{1.00}}$       & $59.69_{7.00}^\ast$          & $71.28_{2.60}^\ast$           & $\uline{77.02_{0.84}^\ast}$     & $79.54_{1.42}^\ast$          & $73.53_{1.53}$               & $59.09_{0.92}^\ast$          & $\uline{65.07_{2.69}}$       & $69.31_{1.52}^\ast$           & $74.33_{1.42}^\ast$            \\
WSSS-Tissue   \cite{han2022multi}        & $76.58_{1.18}^\ast$          & $62.50_{2.86}^\ast$          & $58.45_{13.19}^\ast$         & $65.84_{5.55}^\ast$           & $70.29_{2.16}^\ast$             & $77.94_{0.76}^\ast$          & $72.34_{0.86}^\ast$          & $\mathbf{61.22_{0.72}}$      & $64.24_{2.16}^\ast$          & $68.93_{0.74}^\ast$           & $73.31_{0.73}^\ast$            \\
OEEM   \cite{li2022online}               & $79.14_{1.89}^\ast$          & $67.73_{4.04}^\ast$          & $63.84_{7.82}$               & $70.24_{2.46}^\ast$           & $74.02_{2.61}^\ast$             & $78.95_{3.32}^\ast$          & $73.64_{1.83}$               & $59.63_{4.08}$               & $64.54_{2.46}$               & $69.19_{2.80}^\ast$           & $74.16_{2.67}$                 \\
HAMIL   \cite{zhonghamil}                & $75.59_{0.94}^\ast$          & $60.36_{2.45}^\ast$          & $51.24_{3.28}^\ast$          & $62.40_{2.17}^\ast$           & $68.65_{1.61}^\ast$             & $73.68_{0.47}^\ast$          & $69.64_{0.12}^\ast$          & $57.61_{1.08}^\ast$          & $57.58_{1.65}^\ast$          & $64.63_{0.62}^\ast$           & $69.69_{0.23}^\ast$            \\ \midrule
Ours   (Conf.) \cite{fang2023weakly}     & $80.00_{1.29}^\ast$          & $69.42_{2.87}^\ast$          & $\mathbf{74.07_{2.39}}$      & $\uline{74.50_{1.91}^\ast}$   & $75.50_{1.93}^\ast$             & $\uline{80.91_{0.64}}$       & $\uline{73.92_{1.39}}$       & $59.53_{2.23}$               & $64.36_{1.18}^\ast$          & $\uline{69.68_{0.49}^\ast}$   & $\uline{75.10_{0.64}}$         \\
Ours                                     & $\mathbf{82.17_{0.26}}$      & $\mathbf{74.69_{0.38}}$      & $\uline{73.11_{2.86}}$       & $\mathbf{76.66_{0.98}}$       & $\mathbf{78.84_{0.28}}$         & $\mathbf{81.78_{0.43}}$      & $\mathbf{74.51_{0.64}}$      & $\uline{60.36_{0.45}}$       & $\mathbf{67.25_{0.88}}$      & $\mathbf{70.97_{0.29}}$       & $\mathbf{75.93_{0.41}}$        \\ \bottomrule
\end{tabular}
\end{table*}

\begin{table}[!t]
\setlength{\tabcolsep}{1pt}
\centering
\caption{Performance comparison over the LUAD-HistoSeg dataset. We bold the best and underline the second-best results.}
\label{tab:4}
\resizebox{\columnwidth}{!}{%
\begin{tabular}{@{}cllllll@{}}
\toprule
\multirow{2}{*}{Method}                  & \multicolumn{4}{c}{Tissue IoU (\%)}                                                                         & \multicolumn{1}{c}{\multirow{2}{*}{\begin{tabular}[c]{@{}c@{}}mIoU\\ (\%)\end{tabular}}} & \multicolumn{1}{c}{\multirow{2}{*}{\begin{tabular}[c]{@{}c@{}}fwIoU\\ (\%)\end{tabular}}} \\ \cmidrule(lr){2-5}
                                         & \multicolumn{1}{c}{TE}  & \multicolumn{1}{c}{NEC} & \multicolumn{1}{c}{LYM}       & \multicolumn{1}{c}{TAS} & \multicolumn{1}{c}{}                                                                     & \multicolumn{1}{c}{}                                                                      \\ \midrule 
GradCAM   \cite{8237336}                 & $28.34_{7.96}^{\ast}$   & $0.71_{0.58}^{\ast}$    & $9.06_{7.05}^{\ast}$          & $22.33_{4.66}^{\ast}$   & $15.11_{2.68}^{\ast}$                                                                    & $21.47_{3.13}^{\ast}$                                                                     \\
HistoSegNet   \cite{chan2019histosegnet} & $19.22_{4.28}^{\ast}$   & $20.53_{9.68}^{\ast}$   & $44.43_{2.80}^{\ast}$         & $44.49_{1.70}^{\ast}$   & $32.17_{2.51}^{\ast}$                                                                    & $32.07_{1.58}^{\ast}$                                                                     \\
SEAM   \cite{wang2020self}               & $60.23_{2.36}^{\ast}$   & $54.88_{3.70}^{\ast}$   & $57.85_{3.62}^{\ast}$         & $62.88_{0.94}^{\ast}$   & $58.96_{1.51}^{\ast}$                                                                    & $60.50_{1.55}^{\ast}$                                                                     \\
SC-CAM   \cite{chang2020weakly}          & $72.80_{2.36}^{\ast}$   & $75.04_{5.52}$          & $72.45_{5.30}^{\ast}$         & $68.75_{3.51}^{\ast}$   & $72.26_{4.08}^{\ast}$                                                                    & $71.41_{3.35}^{\ast}$                                                                     \\
WSSS-Tissue   \cite{han2022multi}        & $77.13_{0.48}^{\ast}$   & $\uline{76.26_{1.82}}$  & $\uline{72.71_{0.75}^{\ast}}$ & $\uline{71.42_{0.50}}$  & $\uline{74.38_{0.64}^{\ast}}$                                                            & $\uline{74.36_{0.45}^{\ast}}$                                                             \\
OEEM   \cite{li2022online}               & $72.64_{5.48}^{\ast}$   & $65.34_{10.81}^{\ast}$  & $62.35_{23.69}^{\ast}$        & $67.74_{5.48}^{\ast}$   & $67.02_{11.21}^{\ast}$                                                                   & $68.90_{8.24}^{\ast}$                                                                     \\
HAMIL   \cite{zhonghamil}                & $73.10_{0.96}^{\ast}$   & $63.93_{3.13}^{\ast}$   & $71.39_{1.09}^{\ast}$         & $68.96_{0.51}^{\ast}$   & $69.35_{1.00}^{\ast}$                                                                    & $70.65_{0.68}^{\ast}$                                                                     \\ \midrule
Ours   (Conf.) \cite{fang2023weakly}     & $\uline{78.01_{0.88}}$  & $72.19_{1.79}^{\ast}$   & $70.46_{0.86}^{\ast}$         & $69.75_{1.25}^{\ast}$   & $72.60_{0.37}^{\ast}$                                                                    & $73.50_{0.66}^{\ast}$                                                                     \\
Ours                                     & $\mathbf{78.27_{0.34}}$ & $\mathbf{77.23_{2.05}}$ & $\mathbf{77.00_{0.69}}$       & $\mathbf{72.26_{0.81}}$ & $\mathbf{76.19_{0.66}}$                                                                  & $\mathbf{75.79_{0.45}}$                                                                   \\ \bottomrule
\end{tabular}%
}
\end{table}

\begin{table*}[!t]
\centering
\caption{Performance comparison with different fully-supervised methods over the BCSS dataset. We bold the best and underline the second-best results.}
\label{tab:R16}
\begin{tabular}{@{}ccccccccc@{}}
\toprule
\multirow{2}{*}{\begin{tabular}[c]{@{}c@{}}Back-\\ bone\end{tabular}}  & \multirow{2}{*}{Encoder} & \multicolumn{4}{c}{Tissue IoU (\%)}                                                                                 & \multirow{2}{*}{\begin{tabular}[c]{@{}c@{}}mIoU\\ (\%)\end{tabular}} & \multirow{2}{*}{\begin{tabular}[c]{@{}c@{}}fwIoU\\ (\%)\end{tabular}} & \multirow{2}{*}{\begin{tabular}[c]{@{}c@{}}$\Delta$mIoU\\ (\%)\end{tabular}} \\ \cmidrule(lr){3-6}
                                                                       &                          & TUM                     & STR                            & LYM                            & NEC                     &                                                                      &                                                                       &                                                                              \\ \midrule
\multirow{2}{*}{U-Net}                                                 & ResNet-50                & $80.74_{0.73}^{\ast}$   & $74.95_{0.46}$                 & $\mathbf{63.78_{1.24}^{\ast}}$ & $62.18_{2.23}^{\ast}$   & $70.41_{0.77}$                                                       & $75.81_{0.43}$                                                        & $\mathbf{-0.56}$                                                             \\
                                                                       & EfficientNet-b6          & $81.42_{0.49}$          & $75.12_{0.52}$                 & $62.53_{1.33}^{\ast}$          & $\uline{67.87_{1.46}}$  & $71.73_{0.39}^{\ast}$                                                & $76.28_{0.39}$                                                        & $+0.76$                                                                       \\ \midrule
\multirow{2}{*}{U-Net++}                                                 & ResNet-50                & $79.48_{1.82}^{\ast}$   & $73.90_{1.19}$                 & $63.20_{1.56}^{\ast}$          & $60.42_{3.49}^{\ast}$   & $69.25_{0.93}^{\ast}$                                                & $74.69_{1.24}$                                                        & $\mathbf{-1.72}$                                                             \\
                                                                       & EfficientNet-b6          & $81.22_{0.97}$          & $\uline{75.20_{0.62}}$         & $\uline{63.38_{0.64}^{\ast}}$  & $\mathbf{67.96_{0.71}}$ & $\mathbf{71.94_{0.55}^{\ast}}$                                       & $\uline{76.31_{0.63}}$                                                & $+0.97$                                                                       \\ \midrule
\multirow{2}{*}{\begin{tabular}[c]{@{}c@{}}Deep-\\ LabV3+\end{tabular}} & ResNet-50                & $80.71_{1.09}$          & $74.85_{0.72}$                 & $62.78_{1.09}$                 & $63.09_{2.38}$          & $70.36_{0.55}$                                                       & $75.69_{0.62}$                                                        & $\mathbf{-0.61}$                                                             \\
                                                                       & EfficientNet-b6          & $\uline{81.59_{0.25}}$  & $\mathbf{75.51_{0.31}^{\ast}}$ & $62.94_{1.44}^{\ast}$          & $67.28_{0.97}$          & $\uline{71.83_{0.44}^{\ast}}$                                        & $\mathbf{76.53_{0.18}^{\ast}}$                                        & $+0.86$                                                                       \\ \midrule
\multicolumn{2}{c}{Ours   (Weakly-Supervised)}                                                    & $\mathbf{81.78_{0.43}}$ & $74.51_{0.64}$                 & $60.36_{0.45}$                 & $67.25_{0.88}$          & $70.97_{0.29}$                                                       & $75.93_{0.41}$                                                        & $0.00$                                                                       \\ \bottomrule
\end{tabular}
\end{table*}
The quantitative performance of the proposed HisynSeg framework and other competing methods over the WSSS4LUAD dataset is shown in Table \ref{tab:2}. We also provide the performance of a baseline which shares the same architecture with our HisynSeg framework and is trained utilizing the CAMs generated by a ResNet-50 model with GradCAM \cite{8237336}. It can be seen that the segmentation performance of the proposed HisynSeg not only greatly surpasses the competing methods but also significantly outperforms our previous conference version \cite{fang2023weakly} in all evaluation metrics, except for NOM IoU, where the proposed HisynSeg is the second best. Although our previous work has already achieved SOTA performance in terms of mIoU, HisynSeg further obtains an improvement of over 2\% in mIoU and 3\% in fwIoU, demonstrating the strong ability of the WSSS paradigm based on image-mixing synthesis. 

Table \ref{tab:2} also shows the comparison results over the BCSS-WSSS dataset. Although the proposed HisynSeg framework does not perform the best in single LYM IoU, its overall performance in mIoU and fwIoU remains the best. Besides, compared with our previous work, the new HisynSeg framework improves the performance in all IoUs.

Next, we provide the performance comparison over the LUAD-HistoSeg dataset in Table \ref{tab:4}. This table shows that HisynSeg achieves an mIoU of 76.19\% and an fwIoU of 75.79\%. It is also worth to note that our conference version achieves the second best in mIoU and fwIoU among all the competing methods. However, compared with the conference work, the extended HisynSeg achieves an improvement of 3.59\% and 2.29\% in terms of mIoU and fwIoU. Besides, our framework surpasses WSSS-Tissue, the best-performed competing method, with 1.81\% and 1.43\% in terms of mIoU and fwIoU, respectively.

Overall, from the quantitative results, we can conclude that the performance of the proposed HisynSeg framework is the most robust among the three different datasets compared with other methods. For example, considering mIoU, which best reflects the overall segmentation performance, the best competing method in WSSS4LUAD and BCSS-WSSS is SC-CAM, and the best competing method in LUAD-HistoSeg is WSSS-Tissue. However, the proposed HisynSeg framework steadily achieves the best over all the three datasets, demonstrating its high generalization ability.

To further demonstrate the effectiveness of HisynSeg, we also compare it with models trained under a fully-supervised scheme by utilizing the pixel-level masks in BCSS. Note that the training, validation, and test images are identical to those used in the WSSS setting. The results are listed in Table \ref{tab:R16}. This table shows that our framework based on the WSSS scheme can perform better than all fully-supervised backbones with a ResNet-50 encoder in terms of mIoU and fwIoU. Our approach even shows significantly better performance than U-Net++ with ResNet-50 in mIoU. For other backbones, the proposed HisynSeg has a performance degradation of less than 1\% in terms of mIoU compared with the fully-supervised methods as well. This result indicates that our proposed framework can be a strong counterpart to the fully-supervised methods with fewer annotation efforts.

\begin{figure*}[!t]
    \centering
    \includegraphics[width=0.92\linewidth]{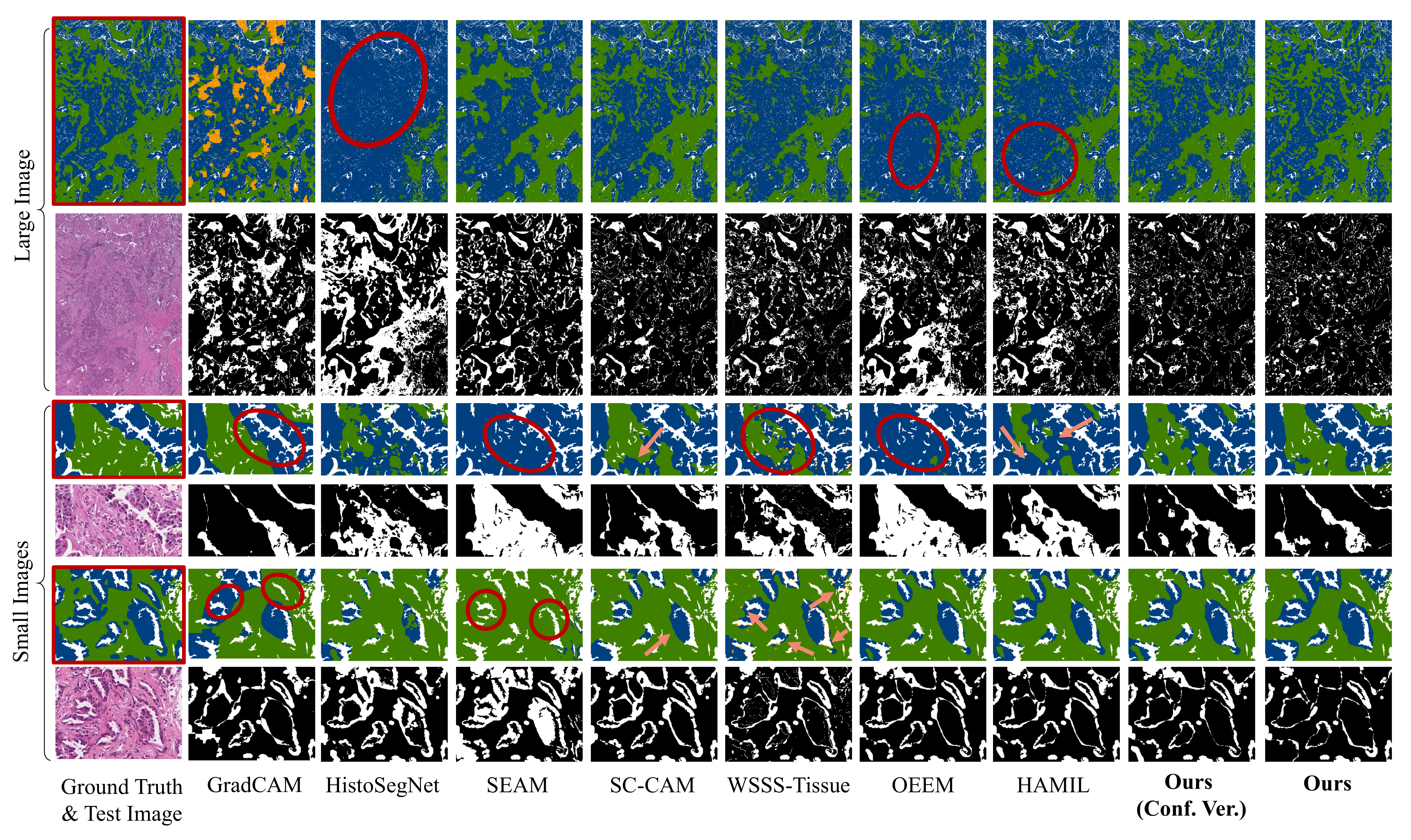}
    \caption{The qualitative illustrations of the segmentation masks of different methods over the WSSS4LUAD dataset. \textcolor[RGB]{0,64,128}{Blue}, \textcolor[RGB]{64,128,0}{green}, and \textcolor[RGB]{243,152,0}{orange} pixels represent tumor, stroma, and normal tissues. The segmentation masks with red boundaries in the first column are the ground truths. The black and white masks represent the differences between the predicted segmentation mask and the ground truth, where black represents true segmentation results, and white stands for wrong segmented pixels. Red circles and orange arrows highlight the wrong segmented pixels.}
    \label{fig:3}
\end{figure*}

\subsubsection{Qualitative Analysis}
To better study the advantages and disadvantages of all the methods, several representative histopathological images and their predicted segmentation masks are selected and illustrated in Fig. \ref{fig:3}. We highlight the wrong predicted pixels with red circles and orange arrows in the figure. From the figure, we can find that the GradCAM's segmentation results in the small images exhibit numerous incorrect tissue boundaries, which is consistent with the issues of under-activation and over-activation in CAM. Moreover, in the result of the large image, GradCAM mistakenly identifies certain tissues as normal. This misclassification may stem from GradCAM's dependence on a classification network for pseudo-mask generation. Consequently, any inaccuracies in the classifier's output can propagate to the pseudo-masks, ultimately degrading the performance of the segmentation network.
Besides, HistoSegNet tends to segment a large region as the same type. For instance, the predicted mask of HistoSegNet highlighted by the red circle in the first row shows that a large stroma area is misclassified as the tumor. Red circles in the segmentation results of SEAM and OEEM highlight that similar problems exist in these two methods. Although SC-CAM develops a sub-category strategy to avoid the over-activation and under-activation issues of CAM, the pixels around the segmentation boundaries still cannot be correctly segmented. Some wrongly predicted tissue categories can be found in the segmentation results of WSSS-Tissue, which mistakenly predicts both stroma and tumor tissue as normal tissue, as highlighted by the red circles and orange arrows. This phenomenon may be attributed to the fact that the progressive dropout attention module in WSSS-Tissue drops the most discriminative features during the training process, causing a decrease of classification ability. Finally, HAMIL tends to misclassify stroma tissues as tumors, as illustrated by the red circles and orange arrows. Besides, compared with our previous work, the extended HisynSeg has more accurate segmentation boundaries, producing the best results overall.

\subsubsection{Evaluation of Synthesized Image Discriminator}
\begin{table}[!t]
\setlength{\tabcolsep}{1pt}
\centering
\caption{Classification performance of the synthesized image discriminator over the three datasets.}
\label{tab:disc-class}
\resizebox{\columnwidth}{!}{%
\begin{tabular}{@{}ccccccc@{}}
\toprule
\multirow{2}{*}{Dataset}                                                  & \multirow{2}{*}{Strategy} & \multicolumn{4}{c}{Metrics (\%)}                                  & \multirow{2}{*}{\begin{tabular}[c]{@{}c@{}} Real Image \#\\ / 10K Images\end{tabular}} \\ \cmidrule(lr){3-6}
                                                                          &                           & Accuracy       & Precision      & Recall         & F1-Score       &                                                                                               \\ \midrule
\multirow{2}{*}{\begin{tabular}[c]{@{}c@{}}WSSS4\\ LUAD\end{tabular}}     & Mosaic                    & $97.44_{1.64}$ & $97.61_{1.39}$ & $96.89_{2.36}$ & $97.15_{1.87}$ & $27.0^\ast$                                                                                   \\
                                                                          & Bézier                    & $91.03_{5.27}$ & $91.97_{4.32}$ & $89.89_{7.25}$ & $89.75_{6.60}$ & $1608.6$                                                                                      \\ \midrule
\multirow{2}{*}{\begin{tabular}[c]{@{}c@{}}BCSS-\\ WSSS\end{tabular}}     & Mosaic                    & $99.01_{1.11}$ & $99.19_{0.91}$ & $98.82_{1.34}$ & $98.98_{1.15}$ & $193.6$                                                                                       \\
                                                                          & Bézier                    & $98.43_{1.65}$ & $98.53_{1.54}$ & $98.30_{1.83}$ & $98.37_{1.71}$ & $867.6$                                                                                       \\ \midrule
\multirow{2}{*}{\begin{tabular}[c]{@{}c@{}}LUAD-\\ HistoSeg\end{tabular}} & Mosaic                    & $96.74_{3.26}$ & $97.20_{2.48}$ & $96.48_{3.62}$ & $96.65_{3.39}$ & $587.2$                                                                                       \\
                                                                          & Bézier                    & $83.31_{4.27}$ & $84.88_{3.97}$ & $82.79_{3.69}$ & $82.88_{4.21}$ & $2541.0$                                                                                        \\ \midrule
\multicolumn{7}{l}{\begin{tabular}[c]{@{}l@{}}$\ast$ denotes $50,000$ images are synthesized in one run. This is because no image is\\ discriminated as real when synthesizing $10,000$ images.\end{tabular}}                                                            
\end{tabular}%
}
\end{table}
\begin{table}[!t]
\setlength{\tabcolsep}{1pt}
\centering
\caption{Performance of the auxiliary classification head. * means the metric has a significant difference from the others.}
\label{tab:R19}
\resizebox{\columnwidth}{!}{%
\begin{tabular}{@{}ccccccc@{}}
\toprule
Dataset                                                                        & Model     & Accuracy (\%)                          & F1-Score (\%)                 & Precision (\%)                 & Recall (\%)                    & AUROC (\%)                     \\ \midrule
\multirow{2}{*}{\begin{tabular}[c]{@{}c@{}}WSSS4-\\      LUAD\end{tabular}}    & ResNet & $85.85_{0.85}$                          & $79.33_{0.54}$                 & $86.05_{1.28}$                 & $74.32_{1.58}$                 & $89.26_{0.75}$                 \\
                                                                               & Ours      & $\mathbf{92.52_{0.71}^{\ast}}$          & $\mathbf{87.44_{1.54}^{\ast}}$ & $\mathbf{90.54_{1.58}^{\ast}}$ & $\mathbf{84.77_{2.57}^{\ast}}$ & $\mathbf{93.96_{0.90}^{\ast}}$ \\ \midrule
\multirow{2}{*}{\begin{tabular}[c]{@{}c@{}}BCSS-\\      WSSS\end{tabular}}     & ResNet & \textbf{$\mathbf{88.72_{0.39}^{\ast}}$} & $\mathbf{80.53_{0.61}^{\ast}}$ & $84.30_{1.62}$                 & $\mathbf{77.68_{0.35}^{\ast}}$ & $\mathbf{91.51_{0.46}}$        \\
                                                                               & Ours      & $87.45_{0.29}$                          & $78.28_{0.64}$                 & $\mathbf{87.57_{1.26}^{\ast}}$ & $71.04_{1.02}$                 & $90.90_{0.55}$                 \\ \midrule
\multirow{2}{*}{\begin{tabular}[c]{@{}c@{}}LUAD-\\      HistoSeg\end{tabular}} & ResNet & $89.66_{0.27}$                          & $85.23_{0.55}$                 & $96.04_{0.61}$                 & $78.20_{0.95}$                 & $93.22_{0.27}$                 \\
                                                                               & Ours      & $\mathbf{90.94_{0.73}^{\ast}}$          & $\mathbf{90.44_{0.77}^{\ast}}$ & $\mathbf{97.17_{0.90}}$        & $\mathbf{84.71_{1.40}^{\ast}}$ & $\mathbf{96.03_{0.25}^{\ast}}$ \\ \bottomrule
\end{tabular}%
}
\end{table}
To demonstrate the effectiveness of the image filtering module, we synthesize 10,000 images and use the image discriminator to determine which of the 10,000 images are real, and the average number of images classified as real is reported. The results are listed in Table \ref{tab:disc-class}, which provides the micro-averaged accuracy along with the macro-averaged precision, recall, and F1-Score. We can see that except for LUAD-HistoSeg with Bézier mask generation, all the other discriminators achieve an accuracy of over 90\%, demonstrating that the discriminator can learn the differences between the real and synthesized images. 
Overall, compared with Bézier mask generation, fewer images synthesized by Mosaic transformation are identified as real images by the discriminator. This phenomenon indirectly verifies that images synthesized by Bézier mask generation have higher authenticity. The results also prove that even when randomly generated, some synthesized images still exhibit high similarities to real images that the discriminator cannot detect, demonstrating the feasibility of the proposed image filtering module.

\subsubsection{Evaluation of Auxiliary Classification Head}
To further validate the performance of the auxiliary classification head, we conduct experiments over the three datasets. We train a ResNet-50 model as an anchor using image-level labels under a fully-supervised scheme. The results are listed in Table \ref{tab:R19}. Our auxiliary classification head performs even better than the ResNet-50 model tailored for classification in all metrics over WSSS4LUAD and LUAD-HistoSeg datasets. Statistical results show that the differences are significant for almost all metrics. For the BCSS-WSSS dataset, although the auxiliary classification head does not achieve better performance than ResNet-50 overall, our framework still performs better in metrics such as precision compared with the ResNet-50 model. 
The good classification performance of our auxiliary classification head shows that it can potentially guarantee the high quality of the activation map, thus ensuring the effectiveness of the proposed consistency regularization.

\begin{figure}[!t]
    \centering
    \includegraphics[width=0.85\linewidth]{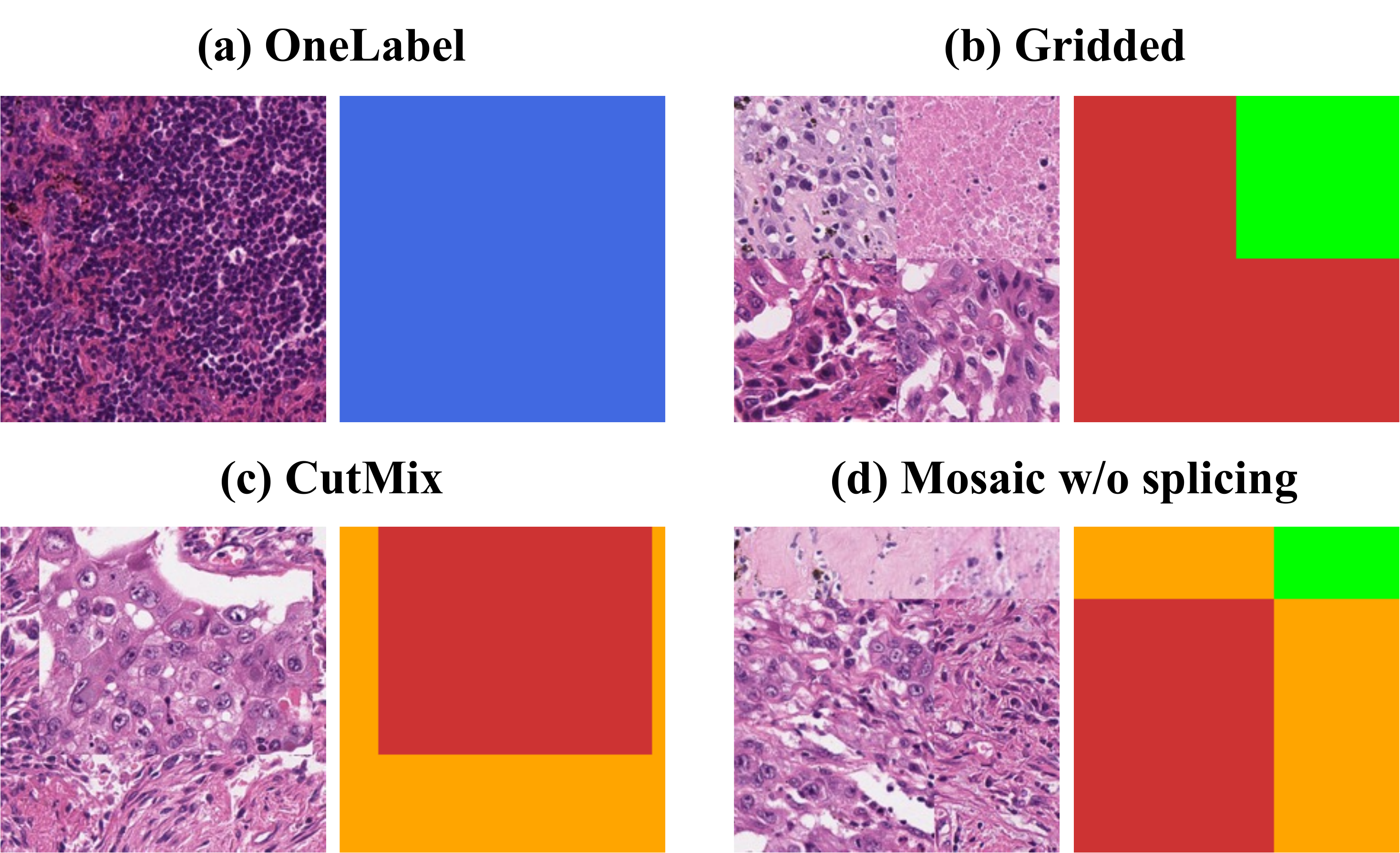}
    \caption{Examples of synthesized images and masks with different strategies. \textcolor[RGB]{205,51,51}{Red}, \textcolor[RGB]{0,255,0}{green}, \textcolor[RGB]{65,105,225}{blue}, and \textcolor[RGB]{255,165,0}{orange} pixels represent tumor epithelial, necrosis, lymphocyte, and tumor-associated stroma tissues.}
    \label{fig:8}
\end{figure}

\subsection{Ablation Studies}
\begin{figure*}[!th]
    \centering
    \includegraphics[width=0.85\linewidth]{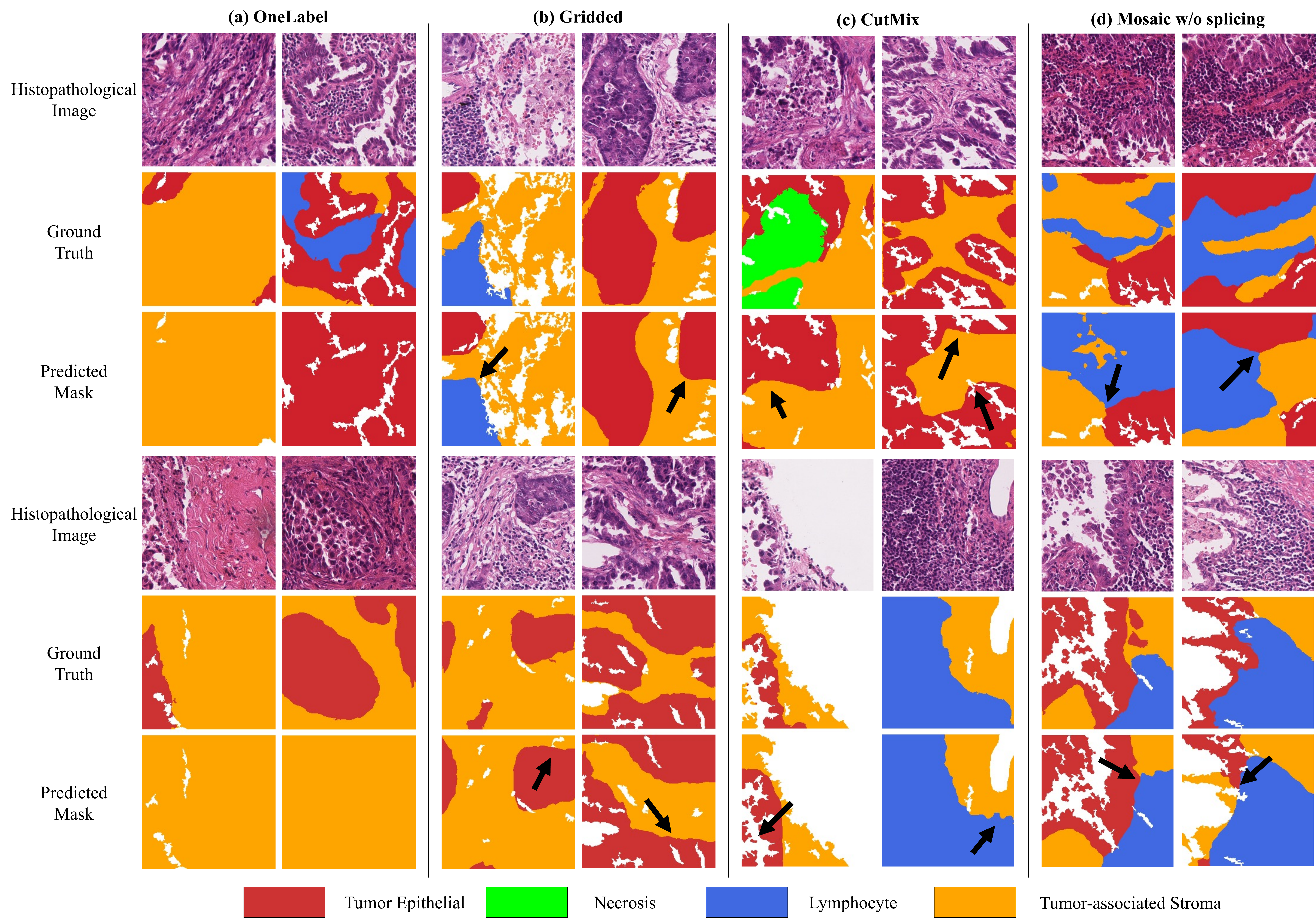}
    \caption{Segmentation results of different synthesis strategies over the LUAD-HistoSeg dataset. Arrows show mispredictions associated with the artifacts in the synthesized images. \textcolor[RGB]{205,51,51}{Red}, \textcolor[RGB]{0,255,0}{green}, \textcolor[RGB]{65,105,225}{blue}, and \textcolor[RGB]{255,165,0}{orange} pixels represent tumor epithelial, necrosis, lymphocyte, and tumor-associated stroma tissues.}
    \label{fig:r1-1}
\end{figure*}

\begin{table}[!t]
\setlength{\tabcolsep}{1pt}
\centering
\caption{Comparison of synthesis strategies over the LUAD-HistoSeg dataset. We bold the best and underline the second-best results.}
\label{tab:R112}
\resizebox{\columnwidth}{!}{%
\begin{tabular}{@{}ccccccc@{}}
\toprule
\multirow{2}{*}{Method}                                       & \multicolumn{4}{c}{Tissue  IoU (\%)}                                                                                    & \multirow{2}{*}{\begin{tabular}[c]{@{}c@{}}mIoU\\ (\%)\end{tabular}} & \multirow{2}{*}{\begin{tabular}[c]{@{}c@{}}fwIoU\\ (\%)\end{tabular}} \\ \cmidrule(lr){2-5}
                                                              & TE                            & NEC                           & LYM                     & TAS                           &                                                                      &                                                                       \\ \midrule
OneLabel                                                      & $69.96_{1.78}^{\ast}$         & $67.81_{2.46}^{\ast}$         & $71.99_{1.32}^{\ast}$   & $66.26_{0.44}^{\ast}$         & $69.00_{0.41}^{\ast}$                                                & $68.70_{0.85}^{\ast}$                                                 \\
Gridded                                                       & $73.67_{0.70}^{\ast}$         & $70.74_{1.05}^{\ast}$         & $\uline{76.05_{1.00}}$  & $69.00_{0.43}^{\ast}$         & $72.37_{0.56}^{\ast}$                                                & $72.04_{0.46}^{\ast}$                                                 \\
Cutmix                                                        & $\uline{75.23_{0.52}^{\ast}}$ & $\uline{70.89_{3.03}^{\ast}}$ & $75.43_{1.46}$          & $\uline{70.86_{0.49}^{\ast}}$ & $\uline{73.10_{1.08}^{\ast}}$                                        & $\uline{73.31_{0.54}^{\ast}}$                                         \\
\begin{tabular}[c]{@{}c@{}}Mosaic w/o\\ splicing\end{tabular} & $74.05_{0.71}$                & $70.42_{2.72}^{\ast}$         & $75.94_{1.89}$          & $69.07_{1.12}^{\ast}$         & $72.37_{1.19}^{\ast}$                                                & $72.18_{0.89}^{\ast}$                                                 \\ \midrule
Ours                                                          & $\mathbf{78.27_{0.34}}$       & $\mathbf{77.23_{2.05}}$       & $\mathbf{77.00_{0.69}}$ & $\mathbf{72.26_{0.81}}$       & $\mathbf{76.19_{0.66}}$                                              & $\mathbf{75.79_{0.45}}$                                               \\ \bottomrule
\end{tabular}%
}
\end{table}

\subsubsection{Comparison of Different Synthesis Strategies}
To show the reasonableness of the image-mixing synthesis module, we study and compare different synthesis strategies, including directly using images with a single tissue type (OneLabel), gridding (Gridded), CutMix \cite{yun2019cutmix}, or using Mosaic transformation without image splicing (Mosaic w/o splicing) over the LUAD-HistoSeg dataset. An example image synthesized by each strategy and the corresponding masks are shown in Fig. \ref{fig:8}. The quantitative results are listed in Table \ref{tab:R112}. To better illustrate the drawbacks of each competing method, typical failure cases are exhibited in Fig. \ref{fig:r1-1}.

The results show that the OneLabel strategy performs poorly because each training image only has one tissue type, misleading the segmentation model. The predicted segmentation masks shown in Fig. \ref{fig:r1-1} also prove that the model trained with the OneLabel strategy tends to segment all pixels as the same tissue type. The poor performance of the Gridded strategy may be attributed to its checkerboard effects, as some of the segmentation boundaries are straight lines or have square corners. In addition, the model trained with CutMix tends to segment all pixels in a rectangular region as the same tissue type. This may be because CutMix utilizes a crop-and-paste strategy to synthesize images, where a rectangle-shaped region with the same tissue type is pasted onto another image, causing unrealistic tissue boundaries in the synthesized images, thereby negatively affecting the model’s performance. In Mosaic w/o splicing, the segmented regions with different types of tissues tend to be rectangles, which corresponds with the transformation process where four different rectangle-shaped images are placed in four corners. To conclude, the proposed synthesis module beats all competing strategies in all types of IoUs, achieving the least improvement of 3.09\% in mIoU, as indicated in Table \ref{tab:R112}.

\begin{table*}[!t]
\setlength{\tabcolsep}{1pt}
\centering
\caption{Comparison results with different semi-supervised methods over WSSS4LUAD and BCSS-WSSS datasets. We bold the best and underline the second-best results.}
\label{tab:R18}
\resizebox{\textwidth}{!}{%
\begin{tabular}{@{}clllll|llllll@{}}
\toprule
\multirow{2}{*}{Method}                                                                     & \multicolumn{5}{c|}{WSSS4LUAD}                                                                                                                                  & \multicolumn{6}{c}{BCSS-WSSS}                                                                                                                                                               \\ \cmidrule(l){2-12} 
                                                                                            & \multicolumn{1}{c}{TUM (\%)}  & \multicolumn{1}{c}{STR (\%)}  & \multicolumn{1}{c}{NOM (\%)}  & \multicolumn{1}{c}{mIoU (\%)} & \multicolumn{1}{c|}{fwIoU (\%)} & \multicolumn{1}{c}{TUM (\%)}  & \multicolumn{1}{c}{STR (\%)} & \multicolumn{1}{c}{LYM (\%)} & \multicolumn{1}{c}{NEC (\%)} & \multicolumn{1}{c}{mIoU (\%)} & \multicolumn{1}{c}{fwIoU (\%)} \\ \midrule
\begin{tabular}[c]{@{}c@{}}CCT   \cite{ouali2020semi}\\      (w/o class label)\end{tabular} & $79.04_{1.39}^{\ast}$         & $66.34_{4.40}^{\ast}$         & $65.28_{3.31}^{\ast}$         & $70.22_{1.69}^{\ast}$         & $73.45_{2.52}^{\ast}$           & $80.01_{0.79}^{\ast}$         & $71.49_{1.79}^{\ast}$        & $55.54_{1.11}^{\ast}$        & $64.21_{1.53}^{\ast}$        & $67.81_{1.04}^{\ast}$         & $73.26_{1.11}^{\ast}$          \\
\begin{tabular}[c]{@{}c@{}}CCT   \cite{ouali2020semi}\\      (w/ class label)\end{tabular}  & $77.83_{3.24}^{\ast}$         & $62.46_{7.81}^{\ast}$         & $\uline{67.81_{3.84}^{\ast}}$ & $69.37_{3.45}^{\ast}$         & $71.25_{4.97}^{\ast}$           & $80.18_{0.65}^{\ast}$         & $71.27_{1.19}^{\ast}$        & $55.41_{1.42}^{\ast}$        & $61.63_{4.95}$               & $67.12_{1.82}^{\ast}$         & $73.13_{0.95}^{\ast}$          \\
CRCFP   \cite{bashir2024consistency}                                                        & $\uline{79.36_{0.52}^{\ast}}$ & $68.38_{1.61}^{\ast}$         & $66.25_{9.99}$                & $71.33_{3.05}^{\ast}$         & $74.49_{0.73}^{\ast}$           & $\uline{80.27_{1.01}^{\ast}}$ & $\uline{73.19_{1.25}}$       & $\uline{59.77_{0.90}}$       & $\mathbf{67.55_{1.57}}$      & $\uline{70.20_{0.80}}$        & $\uline{74.67_{0.75}^{\ast}}$  \\
CPS   \cite{chen2021semi}                                                                   & $61.00_{2.45}^{\ast}$         & $13.49_{8.85}^{\ast}$         & $66.35_{9.47}$                & $46.95_{6.24}^{\ast}$         & $41.75_{5.17}^{\ast}$           & $62.41_{6.29}^{\ast}$         & $47.19_{10.44}^{\ast}$       & $42.83_{10.27}^{\ast}$       & $39.57_{21.59}^{\ast}$       & $48.00_{10.63}^{\ast}$        & $52.98_{8.15}^{\ast}$          \\
UniMatch   \cite{yang2023revisiting}                                                        & $78.81_{1.04}^{\ast}$         & $\uline{69.65_{3.26}^{\ast}}$ & $66.66_{5.12}^{\ast}$         & $\uline{71.71_{2.20}^{\ast}}$ & $\uline{74.71_{1.90}^{\ast}}$   & $78.83_{1.97}^{\ast}$         & $70.31_{1.94}^{\ast}$        & $56.30_{2.07}^{\ast}$        & $64.91_{1.56}$               & $67.59_{1.75}^{\ast}$         & $72.36_{1.86}^{\ast}$          \\ \midrule
Ours                                                                                        & $\mathbf{82.29_{0.39}}$       & $\mathbf{74.40_{1.63}}$       & $\mathbf{73.53_{2.46}}$       & $\mathbf{76.74_{1.31}}$       & $\mathbf{78.81_{0.92}}$         & $\mathbf{81.41_{0.31}}$       & $\mathbf{74.54_{0.42}}$      & $\mathbf{59.93_{0.86}}$      & $\uline{67.18_{1.45}}$       & $\mathbf{70.76_{0.39}}$       & $\mathbf{75.74_{0.23}}$        \\ \bottomrule
\end{tabular}%
}
\end{table*}

\begin{table}[!t]
\setlength{\tabcolsep}{1pt}
\centering
\caption{Comparison of semi-supervised methods over the LUAD-HistoSeg dataset. We bold the best and underline the second-best results. CLS means image-level classification labels.}
\label{tab:19}
\resizebox{\columnwidth}{!}{%
\begin{tabular}{@{}cllllll@{}}
\toprule
\multirow{2}{*}{Method}              & \multicolumn{4}{c}{Tissue IoU (\%)}                                                                                                & \multicolumn{1}{c}{\multirow{2}{*}{\begin{tabular}[c]{@{}c@{}}mIoU\\ (\%)\end{tabular}}} & \multicolumn{1}{c}{\multirow{2}{*}{\begin{tabular}[c]{@{}c@{}}fwIoU\\ (\%)\end{tabular}}} \\ \cmidrule(lr){2-5}
                                     & \multicolumn{1}{c}{TE}   & \multicolumn{1}{c}{NEC}  & \multicolumn{1}{c}{LYM}  & \multicolumn{1}{c}{TAS}  & \multicolumn{1}{c}{}                                                                     & \multicolumn{1}{c}{}                                                                      \\ \midrule
CCT (w/o   CLS) \cite{ouali2020semi} & $73.71_{1.30}^{\ast}$         & $61.45_{3.50}^{\ast}$         & $\uline{73.84_{4.10}^{\ast}}$ & $66.29_{1.74}^{\ast}$         & $68.82_{1.91}^{\ast}$                                                                    & $70.06_{1.34}^{\ast}$                                                                     \\
CCT  (w/ CLS) \cite{ouali2020semi}   & $\uline{74.98_{0.56}^{\ast}}$ & $65.73_{2.61}^{\ast}$         & $72.48_{3.58}^{\ast}$         & $65.34_{1.62}^{\ast}$         & $69.63_{1.59}^{\ast}$                                                                    & $70.38_{1.25}^{\ast}$                                                                     \\
CRCFP   \cite{bashir2024consistency} & $74.87_{1.34}^{\ast}$         & $\uline{70.49_{4.80}^{\ast}}$ & $70.54_{2.11}^{\ast}$         & $\uline{68.46_{0.77}^{\ast}}$ & $\uline{71.09_{1.19}^{\ast}}$                                                            & $\uline{71.59_{0.76}^{\ast}}$                                                             \\
CPS   \cite{chen2021semi}            & $64.04_{0.52}^{\ast}$         & $51.25_{4.00}^{\ast}$         & $70.22_{1.07}^{\ast}$         & $48.07_{2.46}^{\ast}$         & $58.39_{0.62}^{\ast}$                                                                    & $57.99_{0.89}^{\ast}$                                                                     \\
UniMatch   \cite{yang2023revisiting} & $70.44_{0.65}^{\ast}$         & $64.18_{2.32}^{\ast}$         & $68.27_{3.18}^{\ast}$         & $61.52_{3.39}^{\ast}$         & $66.10_{1.11}^{\ast}$                                                                    & $66.37_{1.65}^{\ast}$                                                                     \\ \midrule
Ours                                 & $\mathbf{78.82_{0.64}}$       & $\mathbf{77.09_{2.73}}$       & $\mathbf{77.12_{0.65}}$       & $\mathbf{72.94_{0.85}}$       & $\mathbf{76.49_{0.78}}$                                                                  & $\mathbf{76.28_{0.59}}$                                                                   \\ \bottomrule
\end{tabular}%
}
\end{table}
\subsubsection{Comparison of Semi-Supervised Frameworks}
In the proposed HisynSeg, both synthesized images with pixel-level annotations and real images without masks are utilized for segmentation model training with the proposed consistency regularization. To further verify the effectiveness of the consistency regularization, we employ the same synthesized images used to train HisynSeg as fully-supervised samples and real images as unsupervised samples to compare our framework with semi-supervised segmentation frameworks.
Four different frameworks are selected for comparison, where CCT \cite{ouali2020semi}, CPS \cite{chen2021semi}, and UniMatch \cite{yang2023revisiting} are previous SOTA frameworks developed for general semi-supervised segmentation tasks, while CRCFP \cite{bashir2024consistency} is a latest SOTA framework specifically designed for histopathological image segmentation. In addition, we compare two variants of CCT by considering whether image-level classification labels are utilized. The comparison results are shown in Table \ref{tab:R18} and \ref{tab:19}. From the results, we can see that the proposed framework with the consistency regularization can beat all the competing semi-supervised SOTA frameworks in almost all types of IoUs, except for NEC IoU in BCSS-WSSS, where the HisynSeg achieves the second best but shows no significant difference to the best result. The poor performance of semi-supervised SOTA methods may be because they are based on the consistency of segmentation results of images with different perturbations, which cannot provide sufficient supervision for the segmentation backbone. In contrast, our proposed HisynSeg directly supervises the segmentation results of the unlabeled images with the activation maps, which is very relevant to the segmentation task.

\begin{table}[!t]
\setlength{\tabcolsep}{1pt}
\centering
\caption{Performance with limited number of single-label images over the LUAD-HistoSeg dataset. We bold the best and underline the second-best results.}
\label{tab:R114}
\resizebox{\columnwidth}{!}{%
\begin{tabular}{@{}ccccccc@{}}
\toprule
\multirow{2}{*}{\begin{tabular}[c]{@{}c@{}}Number of\\ Images\end{tabular}} & \multicolumn{4}{c}{Tissue IoU (\%)}                                                                        & \multirow{2}{*}{mIoU (\%)} & \multirow{2}{*}{fwIoU (\%)} \\ \cmidrule(lr){2-5}
                                  & TE                 & NEC              & LYM               & TAS                &                            &                             \\ \midrule
$N=10$                            & $77.40_{0.68}$          & $75.67_{2.04}$          & $\uline{76.83_{0.19}}$  & $70.99_{0.70}^{\ast}$   & $\uline{75.22_{0.69}}$     & $74.81_{0.56}^{\ast}$       \\
$N=20$                            & $\uline{77.96_{0.85}}$  & $76.15_{2.36}$          & $75.51_{1.36}$          & $71.19_{0.56}$          & $75.20_{0.85}$             & $\uline{74.99_{0.60}}$      \\
$N=30$                            & $76.83_{0.39}^{\ast}$   & $\uline{76.27_{1.37}}$  & $74.97_{1.79}$          & $\uline{71.41_{0.57}}$  & $74.87_{0.38}^{\ast}$      & $74.53_{0.31}^{\ast}$       \\
$N=40$                            & $77.51_{0.57}$          & $74.49_{2.82}$          & $76.18_{0.85}$          & $71.15_{1.33}$          & $74.83_{0.36}^{\ast}$      & $74.75_{0.54}$              \\ \midrule
Ours                              & $\mathbf{78.27_{0.34}}$ & $\mathbf{77.23_{2.05}}$ & $\mathbf{77.00_{0.69}}$ & $\mathbf{72.26_{0.81}}$ & $\mathbf{76.19_{0.66}}$    & $\mathbf{75.79_{0.45}}$     \\ \bottomrule
\end{tabular}%
}
\end{table}

\subsubsection{Performance with Limited Number of Images with a Single Tissue Type}
Since HisynSeg requires images with a single tissue type to synthesize images with pixel-level masks, the number of images with a single tissue type may impact the performance of the framework. To prove the robustness of our model, we set the number of single-label images for each tissue category to 10, 20, 30, and 40 and train different models over the LUAD-HistoSeg dataset to study whether the performance will drop sharply. 
It should be noted that we only change the number of single-label images for synthesis while keeping the number of synthesized images discriminated as ``real'' for segmentation model training unchanged, ensuring a fair comparison. The results are listed in Table \ref{tab:R114}. In general, mIoU does not drop drastically when there are limited number of images with a single tissue type. The biggest performance drop of mIoU is less than 1.5\% and is still better than the performance of the previous SOTA framework (WSSS-Tissue with $\text{mIoU}=74.38\%$). It is worth noting that the mIoUs for $N=10$ and $N=20$ have no significant differences compared with the model trained with full data. Besides, we can discover no large performance differences when the number of single-label images is changed slightly (i.e., from 10 to 40). 
The results show that even with a limited number of images with a single tissue type, nearly infinite images can still be synthesized due to the randomness of image selection, preprocessing, splicing, and the choice of control points or anchor points. These synthesized images with rich appearance can provide diverse data for segmentation model training, thus achieving satisfactory segmentation performance.

\subsubsection{Quality Analysis of the Synthesized Images}
\begin{table}[!t]
\setlength{\tabcolsep}{1pt}
\centering
\caption{FIDs and KIDs of different synthesized images. $^\ast$ denotes p-values calculated by comparing different synthesis strategies with the same filtering scheme. $^\dagger$ represents p-values calculated by comparing metrics with and without filtering while fixing the synthesis strategy.}
\label{tab:r13}
\resizebox{\columnwidth}{!}{%
\begin{tabular}{@{}cccc|cc|cc@{}}
\toprule
\multicolumn{2}{c}{\multirow{2}{*}{Dataset}}                                               & \multicolumn{2}{c|}{Without Filtering}        & \multicolumn{2}{c|}{With Filtering}           & \multicolumn{2}{c}{P-value$^\dagger$} \\ \cmidrule(l){3-8} 
\multicolumn{2}{c}{}                                                                       & FID$\downarrow$  & 100$\times$KID$\downarrow$ & FID$\downarrow$  & 100$\times$KID$\downarrow$ & FID               & 100$\times$KID               \\ \midrule
\multirow{3}{*}{\begin{tabular}[c]{@{}c@{}}WSSS4\\ LUAD\end{tabular}}     & Mosaic         & $107.7_{0.5316}$ & $13.38_{0.08}$             & $82.56_{12.61}$  & $7.51_{4.05}$              & P\textless{}0.01  & P\textless{}0.01  \\
                                                                          & Bézier         & $55.93_{0.9071}$ & $5.21_{0.14}$              & $45.63_{5.505}$  & $3.82_{0.72}$              & P\textless{}0.01  & P\textless{}0.01  \\
                                                                          & P-value$^\ast$ & P\textless{}0.01 & P\textless{}0.01           & P\textless{}0.01 & P=0.1190                   & -                 & -                 \\ \midrule
\multirow{3}{*}{\begin{tabular}[c]{@{}c@{}}BCSS-\\ WSSS\end{tabular}}     & Mosaic         & $210.2_{0.4670}$ & $27.13_{0.08}$             & $186.5_{4.302}$  & $24.25_{0.66}$             & P\textless{}0.01  & P\textless{}0.01  \\
                                                                          & Bézier         & $100.6_{1.068}$  & $10.95_{0.12}$             & $37.69_{8.849}$  & $3.21_{1.07}$              & P\textless{}0.01  & P\textless{}0.01  \\
                                                                          & P-value$^\ast$ & P\textless{}0.01 & P\textless{}0.01           & P\textless{}0.01 & P\textless{}0.01           & -                 & -                 \\ \midrule
\multirow{3}{*}{\begin{tabular}[c]{@{}c@{}}LUAD-\\ HistoSeg\end{tabular}} & Mosaic         & $160.9_{0.2003}$ & $17.79_{0.05}$             & $159.4_{7.959}$  & $17.51_{1.04}$             & P=0.7619  & P=0.6429  \\
                                                                          & Bézier         & $128.0_{0.4886}$ & $11.96_{0.09}$             & $103.0_{2.705}$  & $9.36_{0.36}$              & P\textless{}0.01  & P\textless{}0.01  \\
                                                                          & P-value$^\ast$ & P\textless{}0.01 & P\textless{}0.01           & P\textless{}0.01 & P\textless{}0.01           & -                 & -                 \\ \bottomrule
\end{tabular}%
}
\end{table}

To further validate that the Bézier mask generation strategy can synthesize images with higher similarity to the real images than Mosaic transformation and prove the effectiveness of the image filtering module, experiments are conducted by calculating the Fréchet Inception Distance (FID) and Kernel Inception Distance (KID) between the synthesized images and the real images. 
FID and KID are usually used to compare the similarity between two sets of images, without requiring the synthesized images and real images to be paired \cite{fan2022fast}. 
It is worth to note that these two metrics have been widely applied to histopathological image generation tasks \cite{ozyoruk2022deep}. The lower the FID and KID are, the more similar the synthesized images are to the real images. The results are listed in Table \ref{tab:r13}. For all the three datasets, the FIDs and KIDs of images synthesized by Bézier mask generation are significantly lower than those synthesized by Mosaic transformation, except for the KID over the WSSS4LUAD dataset, which still shows a statistical trend that the Bézier mask generation approach is better. Similar conclusions can be drawn when comparing the synthesized images generated without and with filtering. Compared with the images without filtering, almost all FIDs and KIDs show significant drops after applying the image filtering module. The experimental results validate that our proposed Bézier mask generation strategy and the image filtering module can greatly improve the authenticity of the synthesized images.
\begin{figure*}[!t]
    \centering
    \includegraphics[width=0.82\linewidth]{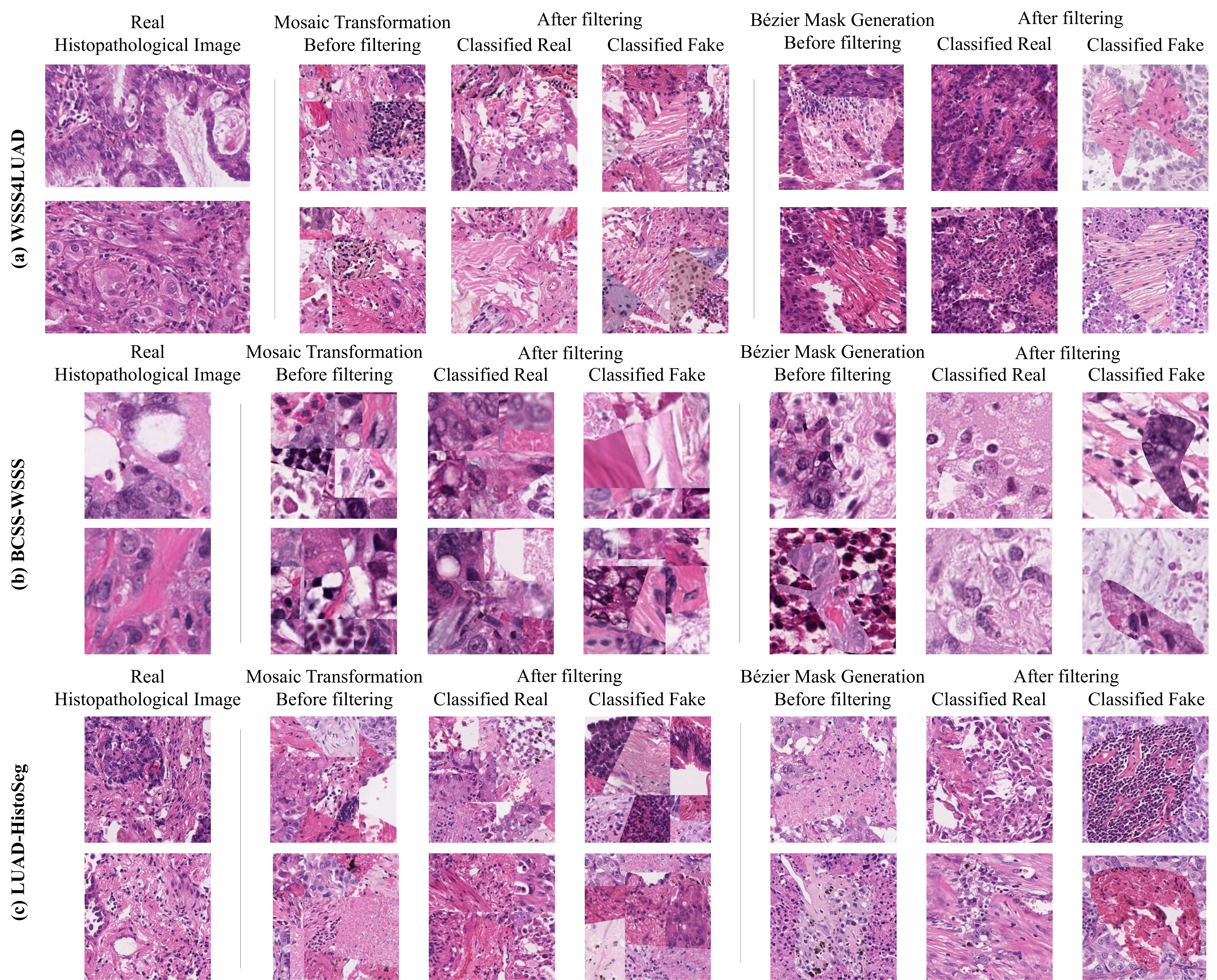}
    \caption{The comparison of real images and synthesized images with Mosaic transformation or Bézier mask generation before and after filtering over WSSS4LUAD, BCSS-WSSS, and LUAD-HistoSeg datasets.}
    \label{fig:r1-2}
\end{figure*}

For better analysis, real and synthesized images are visualized in Fig. \ref{fig:r1-2}. Some synthesized images classified as ``fake'' by the discriminator are also provided. From the figure, it can be seen that although the images without filtering are very close to the real images, there are still some identifiable artifacts due to tumor heterogeneity, such as inconsistent cell density, as shown by the top image generated by Mosaic transformation over the WSSS4LUAD dataset. Furthermore, the images synthesized by Bézier mask generation over the WSSS4LUAD dataset without filtering exhibit inconsistent cell sizes and distortion. These issues likely stem from the varying original resolutions of images within the WSSS4LUAD dataset. Consequently, resizing and synthesizing these images can result in unrealistic outputs.
Differences in stain concentration also cause the synthesized images without filtering to have tissues of various colors, leading to unauthenticity. Similarly, these problems also exist in the images classified as ``fake''. Besides, differences in the sharpness of distinct areas within an image can be found in some of the ``fake'' images, such as the bottom image generated by Bézier mask generation over the LUAD-HistoSeg dataset. However, for the images classified as ``real'', we can see that the staining is more uniform, and the cell size and density do not change sharply, making them more consistent with the real images. These results prove that issues such as various original resolutions of images can be resolved by the filtering module. The second conclusion is that Bézier mask generation consistently synthesizes better images with fewer stitching lines and more even color distributions than Mosaic transformation, proving the effectiveness of the extended Bézier mask generation strategy compared with our conference work.
\begin{figure}
    \centering
    \includegraphics[width=0.9\linewidth]{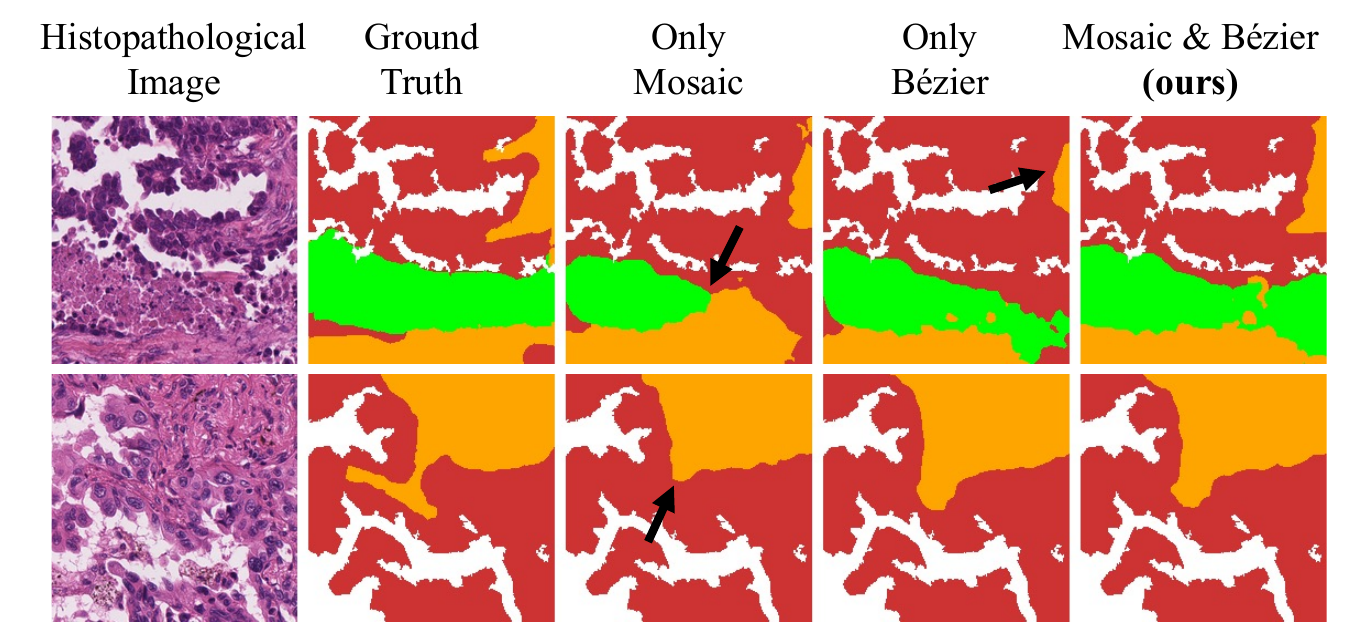}
    \caption{Visualization of predicted masks with separate synthesis strategies over the LUAD-HistoSeg dataset. The black arrows highlight the disadvantages of each strategy. \textcolor[RGB]{205,51,51}{Red}, \textcolor[RGB]{0,255,0}{green}, and \textcolor[RGB]{255,165,0}{orange} pixels represent tumor epithelial, necrosis, and tumor-associated stroma tissues.}
    \label{fig:r2-9}
\end{figure}

\subsubsection{Ablation Studies on Using Separate Synthesis Strategies}
\begin{table}[!t]
\setlength{\tabcolsep}{1pt}
\centering
\caption{Ablation studies using separate synthesis strategies over the LUAD-HistoSeg dataset. We bold the best and underline the second-best results.}
\label{tab:12}
\resizebox{\columnwidth}{!}{%
\begin{tabular}{@{}cccccccc@{}}
\toprule
\multirow{2}{*}{Mosaic} & \multirow{2}{*}{Bézier} & \multicolumn{4}{c}{Tissue IoU (\%)}                                                                        & \multirow{2}{*}{\begin{tabular}[c]{@{}c@{}}mIoU\\ (\%)\end{tabular}} & \multirow{2}{*}{\begin{tabular}[c]{@{}c@{}}fwIoU\\ (\%)\end{tabular}} \\ \cmidrule(lr){3-6}
                        &                         & TE                 & NEC                & LYM                & TAS                &                                                                      &                                                                       \\ \midrule
\checkmark              & \usym{2717}             & $\uline{77.51_{0.68}}$  & $70.88_{3.00}^{\ast}$   & $\uline{76.39_{1.85}}$  & $\mathbf{72.49_{0.61}}$ & $74.32_{0.91}^{\ast}$                                                & $\uline{75.00_{0.58}}$                                                \\
\usym{2717}             & \checkmark              & $77.27_{0.77}$          & $\uline{75.90_{1.55}}$  & $75.45_{1.55}$          & $70.71_{0.64}^{\ast}$   & $\uline{74.83_{0.37}^{\ast}}$                                        & $74.49_{0.52}^{\ast}$                                                 \\ \midrule
\checkmark              & \checkmark              & $\mathbf{78.27_{0.34}}$ & $\mathbf{77.23_{2.05}}$ & $\mathbf{77.00_{0.69}}$ & $\uline{72.26_{0.81}}$  & $\mathbf{76.19_{0.66}}$                                               & $\mathbf{75.79_{0.45}}$                                                \\ \bottomrule
\end{tabular}%
}
\end{table}
To study the roles of the two proposed synthesis strategies, we conduct ablation studies by separately utilizing the Mosaic transformation or Bézier mask generation strategy alone over the LUAD-HistoSeg dataset. Note that the number of training samples remains the same to ensure fairness. The results are listed in Table \ref{tab:12}. 
The table shows that the TE IoU of using the two synthesis strategies individually is similar to that of using the two strategies simultaneously, with a difference of less than or equal to 1\%. But when both strategies are adopted, the NEC IoU is greatly improved and better overall segmentation performance can be achieved. Besides, the table shows no large performance differences between Mosaic transformation and Bézier mask generation. 

To better illustrate the advantages of different synthesis strategies, some predicted masks are also illustrated in Fig.~\ref{fig:r2-9}. The predicted masks using only Mosaic transformation shows that although the segmentation prediction is generally accurate, certain segmentation boundaries exhibit relatively straight curves and near-right-angle turns, deviating from real tissue contours. In contrast, models trained on images synthesized using Bézier mask generation tend to produce smoother tissue boundaries. However, solely utilizing Bézier mask generation sometimes fails to fully delineate the entire tissue areas due to excessively smooth boundaries. The above phenomena indicate that when a single synthesis strategy is utilized, the segmentation model will overfit the artifacts in the synthesized images under that strategy. 
In contrast, when the two synthesis strategies are applied simultaneously, the model is forced to learn the semantic information rather than the artifacts in the synthesized images, since overfitting the specific artifacts of one strategy will decrease the segmentation performance for the images synthesized by the other. Therefore, the overall segmentation performance can be improved, which is validated by the last column of Fig.~\ref{fig:r2-9}.

\begin{table*}
\setlength{\tabcolsep}{1pt}
\centering
\caption{Ablation studies on the synthesized image filtering module over WSSS4LUAD and LUAD-HistoSeg datasets. * means the metric has a significant difference from the other.}
\label{tab:13}
\begin{tabular}{@{}cccccc|cccccc@{}}
\toprule
\multirow{2}{*}{Filtering} & \multicolumn{5}{c|}{WSSS4LUAD}                                                                                                                                     & \multicolumn{6}{c}{LUAD-HistoSeg}                                                                                                                                           \\ \cmidrule(l){2-12} 
                           & TUM (\%)                       & STR (\%)                       & NOM (\%)                       & mIoU (\%)                      & fwIoU (\%)                     & TE (\%)                & NEC (\%)                       & LYM (\%)                & TAS (\%)                & mIoU (\%)                      & fwIoU (\%)              \\ \midrule
\usym{2717}                & $80.84_{0.96}$                 & $71.77_{2.01}$                 & $54.51_{12.97}$                & $69.04_{4.54}$                 & $76.35_{1.45}$                 & $\mathbf{78.61_{0.45}}$ & $73.78_{1.92}$                 & $74.85_{2.10}$          & $\mathbf{73.11_{0.54}}$ & $75.09_{0.68}$                 & $75.71_{0.39}$          \\
\checkmark                 & $\mathbf{82.17_{0.26}^{\ast}}$ & $\mathbf{74.69_{0.38}^{\ast}}$ & $\mathbf{73.11_{2.86}^{\ast}}$ & $\mathbf{76.66_{0.98}^{\ast}}$ & $\mathbf{78.84_{0.28}^{\ast}}$ & $78.27_{0.34}$          & $\mathbf{77.23_{2.05}^{\ast}}$ & $\mathbf{77.00_{0.69}}$ & $72.26_{0.81}$          & $\mathbf{76.19_{0.66}^{\ast}}$ & $\mathbf{75.79_{0.45}}$ \\ \bottomrule
\end{tabular}
\end{table*}
\subsubsection{Ablation Studies on Image Filtering Module}
To validate the synthesized image filtering module, we train a segmentation model with synthesized images that do not pass the filtering module while keeping other settings the same over both WSSS4LUAD and LUAD-HistoSeg datasets. Experimental results in Table \ref{tab:13} show that the segmentation model trained with the filtering strategy outperforms that without the module in both mIoU and fwIoU over the two datasets. 
Compared with the LUAD-HistoSeg dataset, the filtering module exhibits more pronounced performance improvements when applied to WSSS4LUAD. This disparity can be attributed to the homogeneous nature of LUAD-HistoSeg, where images originate from a single source and share a consistent original resolution. Therefore, the synthesized images remain relatively realistic even without filtering.
Conversely, WSSS4LUAD encompasses images from two distinct cohorts with varying original resolutions. Directly resizing these diverse images for synthesis can result in distortion and unreality. Hence, the filtering module's ability to eliminate these unrealistic images significantly enhances the performance of the framework. This demonstrates the filtering module's capacity to improve the framework's robustness to data quality variations. The proposed HisynSeg can achieve impressive segmentation performance even when faced with substantial differences in image appearance.

\begin{table}[!t]
\setlength{\tabcolsep}{1pt}
\centering
\caption{Ablation studies on loss combinations over the LUAD-HistoSeg dataset. We keep the synthesized images the same over the five runs to study the effects of loss combinations. We bold the best and underline the second-best results.}
\label{tab:14}
\resizebox{\columnwidth}{!}{%
\begin{tabular}{@{}ccccccc@{}}
\toprule
\multirow{2}{*}{Settings}                                                           & \multicolumn{4}{c}{Tissue IoU (\%)}                                                                                    & \multicolumn{1}{c}{\multirow{2}{*}{mIoU (\%)}} & \multicolumn{1}{c}{\multirow{2}{*}{fwIoU (\%)}} \\ \cmidrule(lr){2-5}
                                                                                    & \multicolumn{1}{c}{TE}        & \multicolumn{1}{c}{NEC} & \multicolumn{1}{c}{LYM} & \multicolumn{1}{c}{TAS}       & \multicolumn{1}{c}{}                      & \multicolumn{1}{c}{}                       \\ \midrule
w/o $L_{\text{cls}}$                                                                       & $73.33_{1.16}^{\ast}$         & $62.78_{5.24}^{\ast}$   & $\uline{75.67_{1.24}}$  & $66.03_{1.93}^{\ast}$         & $69.46_{1.52}^{\ast}$                     & $70.15_{1.21}^{\ast}$                      \\
w/o   $L_{\text{reg}}$                                                                     & $\uline{77.62_{0.51}^{\ast}}$ & $\uline{74.39_{2.83}}$  & $73.98_{1.64}^{\ast}$   & $\uline{70.89_{0.93}^{\ast}}$ & $\uline{74.22_{1.20}^{\ast}}$             & $\uline{74.40_{0.81}^{\ast}}$              \\
only   $L_{\text{seg}}$                                                                    & $77.41_{0.89}^{\ast}$         & $73.30_{3.57}$          & $74.89_{1.83}^{\ast}$   & $69.50_{1.05}^{\ast}$         & $73.78_{1.43}^{\ast}$                     & $73.83_{0.95}^{\ast}$                      \\ \midrule
\begin{tabular}[c]{@{}c@{}}$L_{\text{seg}}$+$L_{\text{reg}}$+$L_{\text{cls}}$\\      (ours)\end{tabular} & $\mathbf{78.82_{0.64}}$       & $\mathbf{77.09_{2.73}}$ & $\mathbf{77.12_{0.65}}$ & $\mathbf{72.94_{0.85}}$       & $\mathbf{76.49_{0.78}}$                   & $\mathbf{76.28_{0.59}}$                    \\ \bottomrule
\end{tabular}%
}
\end{table}

\begin{figure*}[!ht]
    \centering
\includegraphics[width=0.85\linewidth]{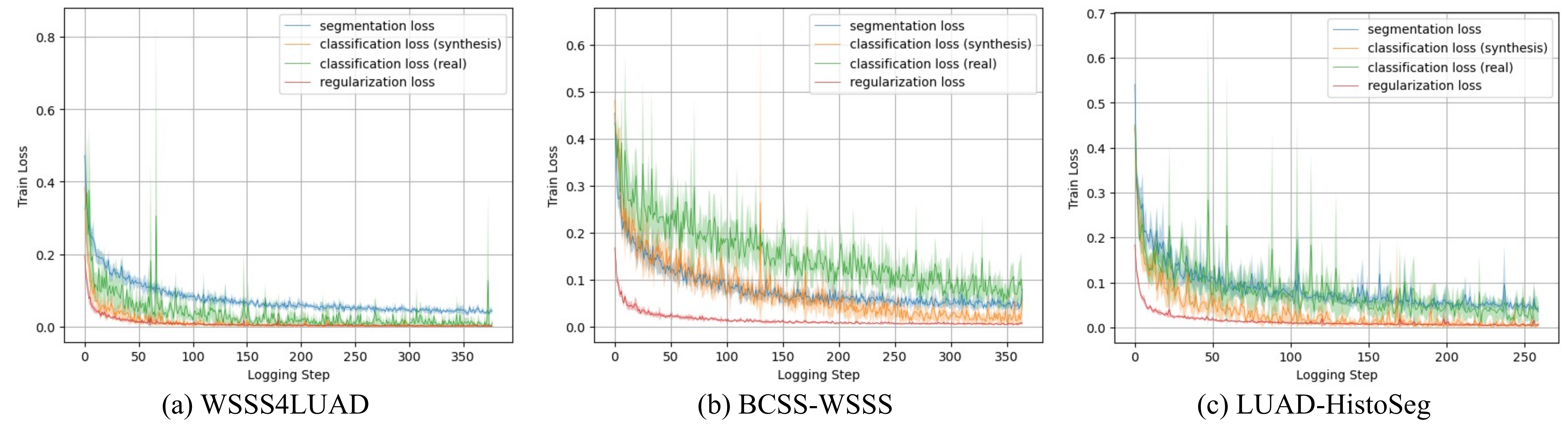}
    \caption{The training loss curves of HisynSeg for each sub-loss over the three datasets. The solid lines represent the mean loss curves over different runs, and the shadows stand for standard deviations.}
    \label{fig:R1.5}
\end{figure*}

\begin{table*}[!t]
\centering
\caption{Performance comparison with different loss weight combinations over the LUAD-HistoSeg dataset. We keep the synthesized images the same over the five runs to study the effects of loss weights. We bold the best and underline the second-best results.}
\label{tab:15}
\begin{tabular}{@{}ccccccccc@{}}
\toprule
\multicolumn{3}{c}{Loss Weights}                                       & \multicolumn{4}{c}{Tissue IoU (\%)}                                                                                                       & \multirow{2}{*}{\begin{tabular}[c]{@{}c@{}}mIoU\\ (\%)\end{tabular}} & \multirow{2}{*}{\begin{tabular}[c]{@{}c@{}}fwIoU\\ (\%)\end{tabular}} \\ \cmidrule(r){1-7}
$w_1$ ($L_\text{cls}$) & $w_2$ ($L_\text{seg}$) & $w_3$ ($L_\text{reg}$) & TE                               & NEC                              & LYM                              & TAS                              &                                                                      &                                                                       \\ \midrule
\multirow{4}{*}{1}     & \multirow{4}{*}{1}     & 0.1                    & $78.61_{0.28}$                   & $75.18_{1.96}$                   & $74.46_{1.78}^{\ast}$            & $72.33_{0.54}$                   & $75.14_{0.61}^{\ast}$                                                & $75.47_{0.49}$                                                        \\
                       &                        & 0.5                    & $\uline{78.64_{0.40}}$     & $73.78_{3.08}$                   & $76.29_{3.05}$                   & $72.34_{0.40}$                   & $75.26_{1.12}$                                                       & $75.63_{0.54}$                                                        \\
                       &                        & 5                      & $72.39_{3.98}^{\ast}$            & $65.55_{5.49}^{\ast}$            & $74.49_{2.37}$                   & $65.12_{6.71}^{\ast}$            & $69.39_{3.80}^{\ast}$                                                & $69.47_{4.40}^{\ast}$                                                 \\
                       &                        & 10                     & $71.67_{1.22}^{\ast}$            & $65.33_{4.57}^{\ast}$            & $71.55_{4.43}^{\ast}$            & $66.66_{3.94}^{\ast}$            & $68.80_{1.90}^{\ast}$                                                & $69.32_{1.61}^{\ast}$                                                 \\ \midrule
\multirow{4}{*}{1}     & 0.1                    & \multirow{4}{*}{1}     & $73.33_{3.30}^{\ast}$            & $71.72_{4.50}$                   & $75.36_{1.78}$                   & $66.77_{6.43}^{\ast}$            & $71.79_{3.31}^{\ast}$                                                & $71.05_{3.97}^{\ast}$                                                 \\
                       & 0.5                    &                        & $78.30_{0.43}$                   & $75.56_{3.11}$                   & \textbf{$\mathbf{77.58_{0.73}}$} &  $\uline{72.48_{0.46}}$    &  $\uline{75.98_{1.13}}$                                         & $\uline{75.84_{0.64}}$                                         \\
                       & 5                      &                        & $78.09_{0.45}$                   & $73.88_{1.44}$                   & $76.08_{0.67}$                   & $71.17_{0.71}^{\ast}$            & $74.80_{0.61}^{\ast}$                                                & $74.94_{0.52}^{\ast}$                                                 \\
                       & 10                     &                        & $77.29_{0.99}^{\ast}$            & $74.20_{4.95}$                   & $75.98_{1.64}$                   & $69.41_{1.35}^{\ast}$            & $74.22_{1.92}^{\ast}$                                                & $73.96_{1.33}^{\ast}$                                                 \\ \midrule
0.1                    & \multirow{4}{*}{1}     & \multirow{4}{*}{1}     & $75.11_{0.30}^{\ast}$            & $67.42_{1.54}^{\ast}$            & $75.97_{1.34}$                   & $67.94_{0.67}^{\ast}$            & $71.61_{0.43}^{\ast}$                                                & $71.99_{0.36}^{\ast}$                                                 \\
0.5                    &                        &                        & $77.58_{0.51}^{\ast}$            & $76.30_{2.08}$                   & $76.85_{1.59}$                   & $71.64_{0.96}$                   & $75.59_{0.48}$                                                       & $75.18_{0.59}^{\ast}$                                                 \\
5                      &                        &                        & $77.14_{0.73}^{\ast}$            & $\uline{76.27_{1.92}}$    & $75.36_{1.32}$                   & $71.04_{0.74}^{\ast}$            & $74.95_{0.96}^{\ast}$                                                & $74.58_{0.76}^{\ast}$                                                 \\
10                     &                        &                        & $76.49_{0.74}^{\ast}$            & $73.87_{1.68}$                   & $75.82_{2.28}$                   & $70.72_{0.83}^{\ast}$            & $74.22_{0.95}^{\ast}$                                                & $74.07_{0.81}^{\ast}$                                                 \\ \midrule
1                      & 1                      & 1                      & \textbf{$\mathbf{78.82_{0.64}}$} & \textbf{$\mathbf{77.09_{2.73}}$} &  $\uline{77.12_{0.65}}$  & \textbf{$\mathbf{72.94_{0.85}}$} & \textbf{$\mathbf{76.49_{0.78}}$}                                     & \textbf{$\mathbf{76.28_{0.59}}$}                                      \\ \bottomrule
\end{tabular}%
\end{table*}
\subsubsection{Ablation Studies on Loss Combinations}
One of this paper's major contributions is proposing an auxiliary classification task and consistency regularization to enable the images without pixel-level masks to participate in the training of the segmentation model. To validate the loss design, ablation studies are conducted with different loss combinations. The results are shown in Table \ref{tab:14}. As we can see from the table, when not applying the classification loss, the segmentation mIoU is only around 70\%, which may result from the fact that the activation map (i.e., $\mathbf{F}_c$) cannot accurately locate the semantic information in histopathological images without the auxiliary classification task, because classification loss also needs to be calculated in CAM-based methods. The wrong activation map makes the regularization useless or even wrong, leading to inferior performance. Besides, the mIoU without the regularization loss is also not satisfactory, which validates the design of the consistency regularization. Finally, we train the model with only the segmentation loss, and the mIoU results in 73.78\%, which is slightly lower than that not utilizing $L_\text{reg}$, proving that the auxiliary classification task can enhance the segmentation performance. Generally, the performance is significantly better in both mIoU and fwIoU when all three loss items are applied.

\begin{table}[!t]
\setlength{\tabcolsep}{1pt}
\centering
\caption{HisynSeg performance with various backbones over the LUAD-HistoSeg dataset. We bold the best and underline the second-best results. DLabV3+: DeepLabV3+; ResNet: ResNet-34; EffiNet: EfficientNet-b6.}
\label{tab:R15}
\resizebox{\columnwidth}{!}{%
\begin{tabular}{cccccccc}
\hline
\multicolumn{2}{c}{\multirow{2}{*}{Backbone}}                                       & \multicolumn{4}{c}{Tissue IoU (\%)}                                                                                                       & \multirow{2}{*}{mIoU (\%)}       & \multirow{2}{*}{fwIoU (\%)}      \\ \cline{3-6}
\multicolumn{2}{c}{}                                                                & TE                               & NEC                              & LYM                              & TAS                              &                                  &                                  \\ \hline
\multirow{2}{*}{U-Net}   & ResNet                                                   & $76.48_{1.16}^{\ast}$            & $72.48_{4.31}$                   & $73.75_{3.17}$                   & $68.64_{2.13}^{\ast}$            & $72.84_{1.16}^{\ast}$            & $72.91_{1.22}^{\ast}$            \\
                         & EffiNet                                                  & $76.99_{0.59}^{\ast}$            & $75.16_{2.15}$                   &  $\uline{76.53_{1.54}}$         & $70.26_{1.48}$                   & $74.74_{1.20}$                   & $74.30_{1.03}^{\ast}$            \\ \hline
\multirow{2}{*}{U-Net++} & ResNet                                                   & $76.65_{0.80}^{\ast}$            & $71.37_{5.02}^{\ast}$            & $72.31_{3.02}^{\ast}$            & $68.01_{1.50}^{\ast}$            & $72.09_{1.80}^{\ast}$            & $72.47_{0.84}^{\ast}$            \\
                         & EffiNet                                                  &  $\uline{78.18_{0.99}}$         & $76.13_{2.07}$                   & $75.58_{1.83}$                   &  $\uline{71.46_{1.75}}$         & $75.34_{1.15}$                   & $\uline{75.19_{1.15}}$         \\ \hline
\multirow{2}{*}{DLabV3+} & ResNet                                                   & $78.07_{1.23}$                   & $\uline{76.20_{1.35}}$         & $76.47_{0.60}$                   & $71.20_{2.35}$                   & $\uline{75.49_{1.07}}$        & $75.17_{1.43}$                   \\
                         & EffiNet & \textbf{$\mathbf{78.27_{0.34}}$} & \textbf{$\mathbf{77.23_{2.05}}$} & \textbf{$\mathbf{77.00_{0.69}}$} & \textbf{$\mathbf{72.26_{0.81}}$} & \textbf{$\mathbf{76.19_{0.66}}$} & \textbf{$\mathbf{75.79_{0.45}}$} \\ \hline
\end{tabular}%
}
\end{table}

\subsubsection{Ablation Studies on Loss Weights}
In HisynSeg, the total loss is calculated by summing up the segmentation loss, regularization loss, and classification loss, which can be viewed as utilizing a weight of 1 for each sub-loss. To validate whether changing the weights may affect the performance of the model, we conduct additional experiments in this subsection. Firstly, we plot the training curves for all the sub-losses by monitoring the value of each sub-loss every 50 training iterations. As shown in Fig. \ref{fig:R1.5}, all sub-losses share similar magnitudes during the training process, regardless of the specific dataset. This result supports our weight setting where we do not handcraft the loss weights to different values.

To further prove that our model is robust to different loss weights, we change the weight of one sub-loss in $\{0.1, 0.5, 1, 5, 10\}$ while fixing the weights of the other two sub-losses over the LUAD-HistoSeg dataset. The results are listed in Table \ref{tab:15}. The table shows that the default setting can achieve the best performance in all types of IoUs except the LYM IoU, where setting the weight for segmentation loss to 0.5 is the best. However, statistical analysis shows no significant difference between the default and best settings in LYM IoU. Statistical tests also show that the IoUs in several settings have no significant differences from our current setting, proving our framework is robust to different loss weights. Hence, we conveniently utilize a fixed weight of 1 for all the sub-losses.

\subsubsection{Ablation Studies on Different Backbones}
Because the proposed image-mixing synthesis module, filtering module, and consistency regularization do not rely on any specifically designed network structure, the HisynSeg framework can be deployed to arbitrary segmentation backbones. In the experiments, we utilize DeepLabV3+ \cite{chen2018encoder} as the backbone to realize image segmentation because it has been widely employed for histopathological image segmentation \cite{han2022multi,zhonghamil}. To validate this choice, we conduct ablation experiments where different backbones are applied to HisynSeg. The results are shown in Table \ref{tab:R15}, indicating that the proposed HisynSeg can achieve an mIoU of more than 72\% regardless of the backbone used.
Besides, DeepLabV3+ shows the best performance than other backbones, and EfficientNet-b6 can beat ResNet-34 in all backbones. These results also prove that using DeepLabV3+ with EfficientNet-b6 can achieve the best performance in all types of IoUs, validating our setting.

\subsubsection{Applying Synthesized Images to Existing WSSS Frameworks}
Besides providing images with pixel-level masks, the two proposed image-mixing synthesis strategies can also be viewed as new data augmentation strategies. Hence, we conduct a series of experiments to check whether the synthesized images can also help boost the performance of existing WSSS frameworks. Specifically, the synthesized images are utilized as extended data to train the WSSS frameworks. Notably, the pixel-level masks are not used in this experiment. Experimental results in Table \ref{tab:R118} validate that the performance of most WSSS frameworks can be improved with the synthesized images. The results further validate the effectiveness of the image-mixing synthesis module.

\begin{table}[!t]
\setlength{\tabcolsep}{1pt}
\centering
\caption{Performance of the existing WSSS frameworks when trained with synthesized images over the LUAD-HistoSeg dataset. * means the metric has a significant difference from the other in the same method.}
\label{tab:R118}
\resizebox{\columnwidth}{!}{%
\begin{tabular}{@{}cccccccc@{}}
\toprule
\multirow{2}{*}{Method}                                                       & \multirow{2}{*}{\begin{tabular}[c]{@{}c@{}}w/ syn. \\ img.\end{tabular}} & \multicolumn{4}{c}{Tissue IoU (\%)}                                                                                                 & \multirow{2}{*}{\begin{tabular}[c]{@{}c@{}}mIoU\\ (\%)\end{tabular}} & \multirow{2}{*}{\begin{tabular}[c]{@{}c@{}}fwIoU\\ (\%)\end{tabular}} \\ \cmidrule(lr){3-6}
                                                                              &                                                                                & TE                             & NEC                            & LYM                              & TAS                            &                                                                      &                                                                       \\ \midrule
\multirow{2}{*}{\begin{tabular}[c]{@{}c@{}}Histo-\\      SegNet\end{tabular}} & \usym{2717}                                                                    & $19.22_{4.28}$                 & $20.53_{9.68}$                 & $44.43_{2.80}$                   & $44.49_{1.70}$                 & $32.17_{2.51}$                                                       & $32.07_{1.58}$                                                        \\
                                                                              & \checkmark                                                                     & $\mathbf{22.16_{6.30}}$        & $\mathbf{24.91_{12.71}}$       & $\mathbf{48.56_{5.70}}$          & $\mathbf{45.23_{5.08}}$        & $\mathbf{35.22_{4.11}}$                                              & $\mathbf{34.46_{3.20}}$                                               \\ \midrule
\multirow{2}{*}{SEAM}                                                         & \usym{2717}                                                                    & $60.23_{2.36}$                 & $54.88_{3.70}$                 & \textbf{$\mathbf{57.85_{3.62}}$} & $62.88_{0.94}$                 & $58.96_{1.51}$                                                       & $60.50_{1.55}$                                                        \\
                                                                              & \checkmark                                                                     & $\mathbf{68.66_{3.64}^{\ast}}$ & $\mathbf{69.17_{2.45}^{\ast}}$ & \textbf{$57.46_{7.41}$}          & $\mathbf{66.16_{2.73}}$        & $\mathbf{65.36_{3.53}^{\ast}}$                                       & $\mathbf{66.27_{3.44}^{\ast}}$                                        \\ \midrule
\multirow{2}{*}{SC-CAM}                                                       & \usym{2717}                                                                    & $72.80_{2.36}$                 & $75.04_{5.52}$                 & $72.45_{5.30}$                   & $68.75_{3.51}$                 & $72.26_{4.08}$                                                       & $71.41_{3.35}$                                                        \\
                                                                              & \checkmark                                                                     & $\mathbf{77.02_{1.18}^{\ast}}$ & $\mathbf{79.76_{2.58}}$        & $\mathbf{75.40_{0.73}}$          & $\mathbf{72.60_{0.89}^{\ast}}$ & $\mathbf{76.20_{1.05}^{\ast}}$                                       & $\mathbf{75.37_{0.91}^{\ast}}$                                        \\ \midrule
\multirow{2}{*}{\begin{tabular}[c]{@{}c@{}}WSSS-\\      Tissue\end{tabular}}  & \usym{2717}                                                                    & $77.13_{0.48}$                 & $76.26_{1.82}$                 & $72.71_{0.75}$                   & $71.42_{0.50}$                 & $74.38_{0.64}$                                                       & $74.36_{0.45}$                                                        \\
                                                                              & \checkmark                                                                     & $\mathbf{78.10_{0.21}^{\ast}}$ & $\mathbf{77.03_{2.57}}$        & $\mathbf{73.59_{0.66}}$          & $\mathbf{71.79_{0.51}}$        & $\mathbf{75.13_{0.90}}$                                              & $\mathbf{75.08_{0.51}}$                                               \\ \midrule
\multirow{2}{*}{OEEM}                                                         & \usym{2717}                                                                    & $72.64_{5.48}$                 & $65.34_{10.81}$                & $62.35_{23.69}$                  & $67.74_{5.48}$                 & $67.02_{11.21}$                                                      & $68.90_{8.24}$                                                        \\
                                                                              & \checkmark                                                                     & $\mathbf{75.58_{0.93}}$        & $\mathbf{71.24_{5.33}}$        & $\mathbf{75.06_{2.55}}$          & $\mathbf{71.19_{0.33}}$        & $\mathbf{73.27_{2.13}}$                                              & $\mathbf{73.50_{1.30}}$                                               \\ \midrule
\multirow{2}{*}{HAMIL}                                                        & \usym{2717}                                                                    & $\mathbf{73.10_{0.96}}$        & $\mathbf{63.93_{3.13}}$        & $\mathbf{71.39_{1.09}^{\ast}}$   & $\mathbf{68.96_{0.51}}$        & $\mathbf{69.35_{1.00}}$                                              & $\mathbf{70.65_{0.68}}$                                               \\
                                                                              & \checkmark                                                                     & $71.68_{2.42}$                 & $63.74_{6.49}$                 & $67.71_{1.50}$                   & $67.76_{1.19}$                 & $67.72_{2.74}$                                                       & $69.10_{2.06}$                                                        \\ \bottomrule
\end{tabular}%
}
\end{table}
\section{Discussion and Conclusion}
In recent years, WSSS for histopathological images has received extensive attention since there is no need to acquire expensive and laborious pixel-level annotations. Currently, mainstream WSSS methods use CAM to generate pseudo-masks and train a segmentation model. However, the CAM has long been criticized for its over-activation and under-activation issues. To this end, this paper proposes a novel WSSS framework named HisynSeg for histopathological images based on image-mixing synthesis and consistency regularization. Considering the artifacts in the synthesized images, this paper proposes a synthesized image filtering module to guarantee the authenticity of the synthesized images. Besides, since it is hard to ensure that there are no artifacts in the filtered synthesized images, this paper further feeds the real images into the segmentation model to avoid the model overfitting the artifacts. Considering there are no segmentation masks for real images to supervise the model, a self-supervised consistency regularization is further proposed to constrain the segmentation probability maps of the real images, enabling the real images without artifacts to take part in the segmentation model training process.

There are some limitations in this study. Since the image filtering module is independent of the synthesis module, knowledge about real and fake images learned by the filtering module cannot be fed back to the synthesis module, preventing the synthesis process from being guided. If a learnable image synthesis module can be developed to synthesize authentic images and pixel-level masks gradually, the computation resource wastage caused by generating fake images can be reduced. Furthermore, since the proposed framework relies on accurate image-level labels to synthesize images, all tissue types in the image should be known. However, in practice, there are some datasets like DigestPath  \cite{da2022digestpath} whose labeling process is based on the MIL scheme. In these datasets, patches will be labeled with ``tumor'' if they contain tumor tissues. Therefore, adopting HisynSeg may lead to over-segmentation for the tumor region because some normal tissues may be regarded as tumors within the tumor patches. However, such a problem will also negatively affect the performance of existing CAM-based WSSS frameworks. Due to the mismatch of multiple-category images and single-category labels, the performance of the classification network utilized in CAM-based methods will inevitably be degraded \cite{yun2021re}, leading to performance drop of the CAM-based methods. Considering that our image-level annotation scheme can reduce the annotation time from 120 seconds per image for pixel-level annotation to 4 seconds \cite{han2022multi}, our framework will have broad applications for actual clinical scenarios.
In the future work, we will continue to study how to solve the label mismatch problem for WSSS tasks.

In conclusion, this paper proposes a novel WSSS framework named HisynSeg for histopathological images based on image-mixing synthesis and consistency regularization. Experimental results on three datasets validate that the proposed framework can beat the SOTA methods, and ablation studies demonstrate the effectiveness of each module.

{\small
\bibliographystyle{ieeetr}	\bibliography{egbib}
}

\end{document}